\documentclass[a4paper, 10pt]{article}
\usepackage[utf8]{inputenc}
\usepackage{amssymb}
\usepackage{amsmath}
\usepackage{graphicx}
\usepackage{subcaption}
\usepackage{hyperref}
\usepackage{fancyhdr}

\fancypagestyle{mypagestyle}{\fancyhf{}\fancyfoot[C]{\footnotesize{This work was supported by MathWorks and Robocup Federation under grant number 2020-14.\\Authors would also thank to the MathWorks Students Competitions Team, Control Systems Team and Advanced Research and Technology Office, for their help in the development of this project.}}}

\title{Mass Estimation in Manipulation Tasks of Domestic Service Robots using Fault Reconstruction Techniques}
\author{Marco Negrete \and Jesús Savage \and José Avendaño}

\begin{document}
\maketitle
\begin{abstract}
  Manipulation is a key capability in domestic service robots, as can be seen in the rulebooks of last Robocup@Home editions. Currently, object recognition is performed based mostly on visual information. Some robots use also 3D information such as point clouds or laser scans but, to the knowledge of authors, robots don't use physical properties to improve object recognition. Estimation of an object's weight during a manipulation task is something new in the @Home league and such ability can improve performance of domestic service robots. In this work we propose to estimate the weight of the grasped object using Sliding Mode Observers. If we consider the manipulator without load as the nominal system and object's weight as a fault signal, we can estimate such weight by an appropriate filtering of the output error injection term of the sliding mode observer. To implement our proposal we used MATLAB and Simulink Robotics System Toolbox, ROS Toolbox and Simscape. To improve computation time we exported all algorithms to standalone ROS nodes from Simulink models. Tests were performed using two platforms: Justina's left manipulator (a robot developed at Biorobotics Laboratory, UNAM) and Neuronics Katana manipulators. We present results in simulation and discuss the performance of the proposed system and the possible sources of error. Finally we present our conclusions and state the future work. 
\end{abstract}

\section{Introduction}
\thispagestyle{mypagestyle}
Consider the following serving-drinks-test scenario in the Robocup@Home league: a robot is asked to bring beverages and to do so, the robot performs the following common steps: it navigates to the bar, recognizes the beverages (usually cans or tetra packs), calculate the inverse kinematics of its manipulator, grasps the object and returns to the user's location. This is a common situation in @Home tasks, nevertheless, what happens if there are empty cans or tetra packs in the bar? As it is commonly implemented in @Home league, robots are unable to distinguish between empty and full cans, since they commonly rely only in visual information of the beverage containers (see, for example, description papers of last edition winner teams \cite{tdp2019Eindhoven, tdp2019Homer, tdp2019UTS}).
\subsection{Objectives and goals}
The project we propose consists in applying model-based fault reconstruction techniques \cite{ding2013model} to estimate the weight of the object being carried by the robot. If we consider the manipulator without load  as the nominal system, and the weight of the object as an external perturbation causing a faulty behavior, we can apply fault reconstruction techniques to estimate such perturbation, i.e., to estimate object's weight. There are several approaches for model-based fault reconstruction. Residual generation is a common technique where an observer is designed to be sensitive to fault signals. On the other hand, Sliding Mode Observers (SMO) are dynamic estimators designed to be robust against fault signals \cite{shtessel2014sliding}. SMOs also provide the ability to reconstruct the fault signal by filtering the so called output error injection term \cite{alwi2011fault}. With the estimation of the manipulated object mass, we expect to improve the manipulation performance in domestic service robots. 

\subsection{Description of the document}
This technical report is organized as follows: In section \ref{sec:background} we briefly introduce the concept of Sliding Mode Observers (SMO) and we describe the robot Justina's manipulator, the platform we chose to test our proposal. In section \ref{sec:estimation} we explain the theoretical part of our proposal while section \ref{sec:implementation} is dedicated to describe the implementation using a model-based design approach in MATLAB and Simulink. Section \ref{sec:results} describe our results and also, the cases where the proposal can fail. In section \ref{sec:reusability} we show the reusability of our proposal by replicating some results using a simulated Katana manipulator. Finally in section \ref{sec:conclusions} we state our conclusions and sketch the future work. 
\section{Background}
\label{sec:background}
\subsection{Sliding Mode Observers}
Sliding mode observers are disticontinuos observers that have the properties of finite time convergence and accurate tracking of the measured states once the sliding surface is reached. These type of observers can be used to reconstruct faults or disturbance signals through an appropiate filtering of the so-called injection output error. Interested readers can find further information about Sliding Mode Observers in \cite{shtessel2014sliding,davila2005second,fridman2002higher}.

Fault detection and isolation, from the control theory approach, is commonly achieved by the implementation of residual generators. A common way to design such residual generators is through observers that are sensitive to the fault signals. Nevertheless, SMOs inherit the properties of Sliding Modes theory, i.e., SMOs are designed to be robust against disturbances and fault signals and to accurately track the system states. As previously explained, the discontinuos term of the SMO will contain information about the fault signal wich can be reconstructed by obtaining the equivalent output error injection. 

\subsection{Robot Justina's manipulator}
Justina is a domestic service robot built at the Biorobotics Laboratory of the National Autonomous University of Mexico and developed under the ViRBot architecture \cite{savage2008virbot}. This robot and its predecessors have been participating in the Robocup@Home league \cite{wachsmuth2015robocup} since 2006 performing several tasks such as cleaning a table, serving drinks, and several other tasks that humans ask for. 

\begin{figure}
  \centering
  \begin{subfigure}[b]{0.4\textwidth}
    \centering
    \includegraphics[width=\textwidth]{Justina.png}
    \caption{Sensors and actuators.}
    \label{fig:Justina}
  \end{subfigure}
  \hfill
  \begin{subfigure}[b]{0.45\textwidth}
    \centering
    \includegraphics[width=\textwidth]{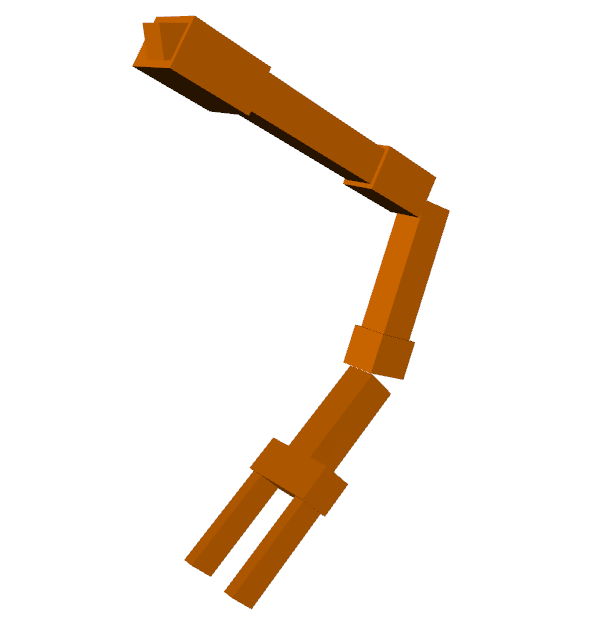}
    \caption{The left manipulator representation for simulation purposes.}
    \label{fig:left_arm}
  \end{subfigure}
  \caption{The domestic service robot Justina.}
\end{figure}

Among other actuators, Justina has two 7-DOF manipulators built with Dynamixel servomotors. Such servomotors allow the user to control them in several forms: by setting a goal position, setting a goal movement speed or setting a desired current which will produce a torque. This latter option is used in this project. Also, these motors provide position sensing with a resolution of less than a tenth of degree. Figure \ref{fig:Justina} shows robot Justina and its sensors and actuators. 

To describe the kinematic chain and some physical properties of the manipulator, we use an URDF file. We already had an URDF describing the whole robot kinematic chain, nevertheless, for this project it was necessary only one of the mainpulators. We chose the left arm and extracted the corresponding tags to a new file. This file was used as a starting point for modeling and simulation purposes. Figure \ref{fig:left_arm} shows the left arm of robot Justina as represented in the URDF.

\section{Mass estimation with Sliding Mode Observers}
\label{sec:estimation}

\subsection{Dynamic Model}
  From the Langrangian of the manipulator, a dynamic model of the following form can be obtained:
  \begin{equation}
    M(q)\ddot{q} + C(q, \dot{q})\dot{q} + B\dot{q} + G(q) + \Delta(q,\dot{q}, u) = u
    \label{eq:lagrangian}
  \end{equation}
  where $q\in \mathbb{R}^7$ are the joint angles, $M(q)\in \mathbb{R}^{7\times 7}$ is the inertia matrix, $C(q,\dot{q})\in \mathbb{R}^{7\times 7}$ is the Matrix of Coriollis forces, $B\dot{q}\in \mathbb{R}^7$ is the vector of friction forces, $G(q)\in\mathbb{R}^7$ is the vector of gravitational forces, $u$ is the input torque, considered as control signal, and $\Delta(q,\dot{q},u)$ is a vector containing all errors due to uncertanties, disturbances and fault signals.
  
  To design a SMO it is necessary to write the model in variable states form. Let $x_1 = [q_1\;q_2\;q_3\;q_4\;q_5\;q_6\;q_7]^T$ and $x_2 = [\dot{q}_1\;\dot{q}_2\;\dot{q}_3\;\dot{q}_4\;\dot{q}_5\;\dot{q}_6\;\dot{q}_7]^T$ be the state variables. Then (\ref{eq:lagrangian}) can be written as:
  \begin{eqnarray}
    \dot{x}_1 &=& x_2\label{eq:model1}\\
    \dot{x}_2 &=& -M^{-1}(q)\left(C(q, \dot{q})\dot{q} + B\dot{q} + G(q) + \Delta(q,\dot{q},u) - u\right)\label{eq:model2}
  \end{eqnarray}
Equation (\ref{eq:model2}) can also be written in the form:
  \begin{equation*}
    \dot{x}_2 = f(x_1, x_2, u) + \phi(x_1, x_2, u)
  \end{equation*}
  where $f(x_1, x_2, u) = -M^{-1}(q)\left(C(q, \dot{q})\dot{q} + B\dot{q} + G(q) - u\right) \in \mathbb{R}^7$ is the nominal part and $\phi(x_1, x_2, u) \in \mathbb{R}^7$ contains all terms related to uncertainties, disturbances and fault signals. If the system is correctly identified, and assuming we have no other disturbances than the object being carried, then $\phi(x_1, x_2, u)$ corresponds only to the fault signals, which, in this work, will be caused by the weight of the object being manipulated.

  Thus, if we reconstruct the signal $\phi$ we will be able to estimate the mass of the manipulated object. 

\subsection{Observer for Disturbance Reconstruction}
If a SMO is used to estimate the joint speeds, the unknown term $\phi(x_1, x_2, u)$ in (\ref{eq:model1})-(\ref{eq:model2}) can be reconstructed by an appropriate filtering of the output error injection term. We used the observer proposed by \cite{shtessel2014sliding}:
  \begin{eqnarray}
    \dot{\hat{x}}_1 &=& \hat{x}_2 + z_1\label{eq:observer1}\\
    \dot{\hat{x}}_2 &=& f(x_1, \hat{x}_2, u) + z_2\label{eq:observer2}
  \end{eqnarray}
  where $z_1$ and $z_2$ are the output error injection terms calculated as
  \begin{equation*}z_1 =
    \left[\begin{tabular}{c}
        $z_{11}$\\
        $\vdots$\\
        $z_{17}$
    \end{tabular}\right] = 
    \left[\begin{tabular}{c}
        $\lambda\vert q_1 - \hat{q}_1\vert ^{1/2}sign(q_1 - \hat{q}_1)$ \\
        $\vdots$\\
        $\lambda\vert q_7 - \hat{q}_7\vert ^{1/2}sign(q_7 - \hat{q}_7)$
    \end{tabular}\right]
\end{equation*}
\begin{equation*}z_2 =
  \left[\begin{tabular}{c}
      $z_{21}$\\
      $\vdots$\\
      $z_{27}$
    \end{tabular}\right] = 
  \left[\begin{tabular}{c}
      $\alpha sign(q_1 - \hat{q}_1)$ \\
      $\vdots$\\
      $\alpha sign(q_7 - \hat{q}_7)$
    \end{tabular}\right]
\end{equation*}

In this observer, the sliding surface is given by $\sigma = x_2 - \hat{x}_2$. When the sliding mode is reached, it holds that:
\[\sigma = \dot{\sigma} = \dot{x}_2 - \dot{\hat{x}}_2 = f(x_1, x_2, u) + \phi(x_1, x_2, u) - f(x_1, \hat{x}_2, u) - z_{2_{eq}} = 0\]
Since,  $x_2 = \hat{x}_2$, then
\begin{equation}
  z_{2_{eq}} = \left[\begin{tabular}{c}
      $z_{21_{eq}}$\\
      $\vdots$\\
      $z_{27_{eq}}$
    \end{tabular}\right] = \phi(x_1, x_2, u) =
  \left[\begin{tabular}{c}
      $\phi_1(q_1,\dots, q_7, \dot{q}_1, \dots, \dot{q}_7, u_1, \dots,u_7)$\\
      $\vdots$\\
      $\phi_7(q_1,\dots, q_7, \dot{q}_1, \dots, \dot{q}_7, u_1, \dots,u_7)$
    \end{tabular}\right]
  \label{eq:zeq}
\end{equation}
where $z_{2_{eq}}$ is the equivalent output error injection which can be obtained by an appropriate low-pass filtering of $z_2$.

Noisy and low sampling frequency rates can cause chattering effects on the estimated states. To mitigate this effect on mass estimation, the output error injection will be low-pass filtered. However, if estimated states are used for a closed-loop control, then actuator signals will have high frequency components that will result in hardware damage. Thus, as it will be later discussed, the SMO is used to estimate the fault signal but, to implement a closed-loop control, an Extended Kalman Filter is used instead. 

\subsection{Mass estimation}
\label{sec:MassEstimation}
The term $\phi(x_1, x_2, u)$ in equation (\ref{eq:model2}) is a function of input torque, joint positions and joint speeds. The mass of the object could be calculated in any point of the state space, nevertheless, calculations are much simpler if such estimation is made only when the manipulator is in a constant position. If $\dot{q}= \ddot{q} = 0$, then $\phi(x_1, x_2, u) = \phi(q)$ depends only on the gravitational torques caused by the weight of the object being carried. From (\ref{eq:model2}) we can see that disturbance $\phi$ is a signal of acceleration, not a torque signal. Let $\tau_o = [\tau_o1,\dots,\tau_o7]^T$ be the torque exerted by the manipulated object with mass $m_o$ and weight $w_o=m_o g$ on each joint. Vector $\tau_o$ can be obtained by multiplying signal $\phi$ by the inertia matrix $M(q)$. Thus, from equation (\ref{eq:zeq}):

\begin{equation}
  \tau_o = M(q)\phi(q) = M(q)z_{2_eq}
  \label{eq:taus}
\end{equation}

\begin{figure}
  \centering
  \begin{subfigure}[b]{0.4\textwidth}
    \centering
    \includegraphics[width=\textwidth]{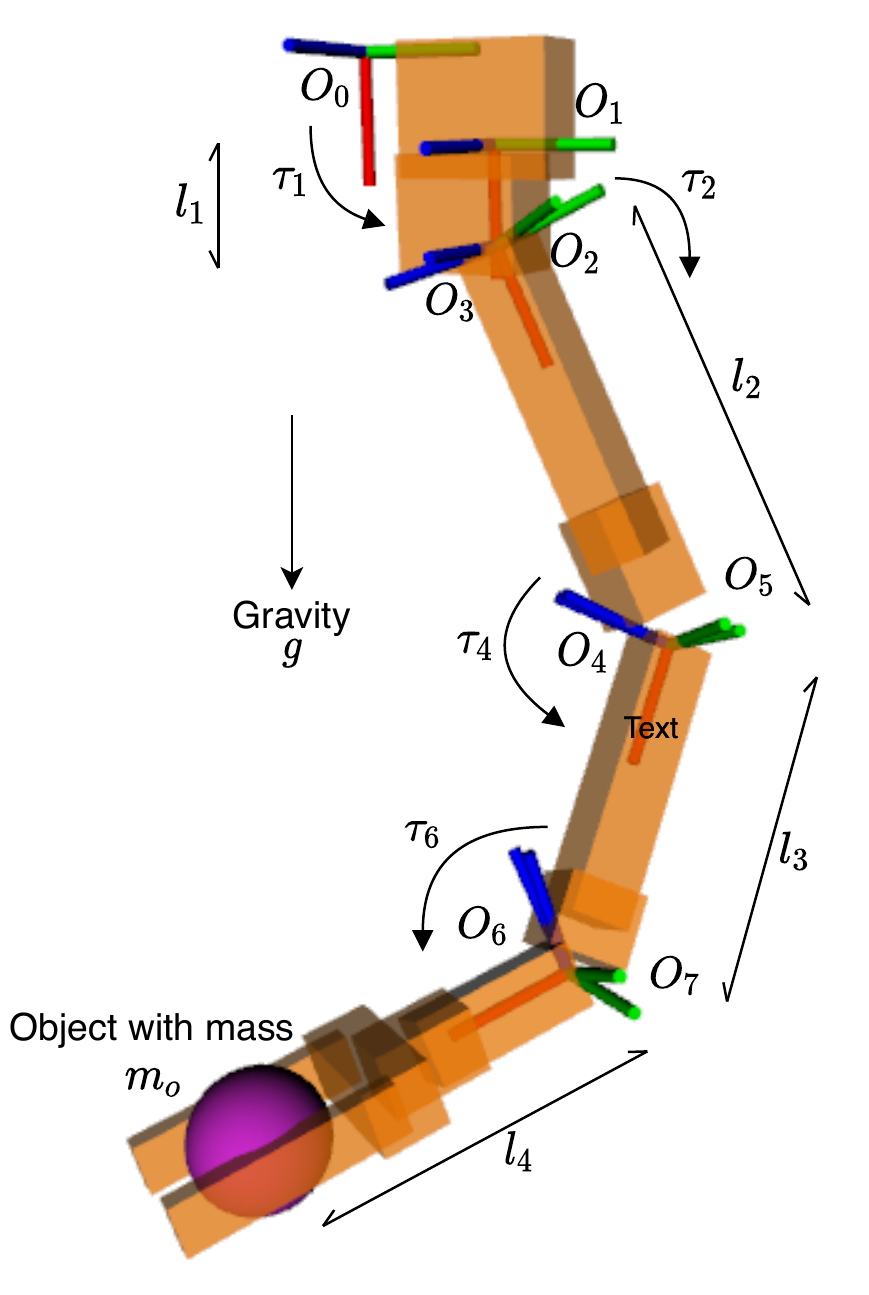}
    \caption{Torques $\tau_o$ on each joint.}
    \label{fig:taus}
  \end{subfigure}
  \hfill
  \begin{subfigure}[b]{0.55\textwidth}
    \centering
    \includegraphics[width=\textwidth]{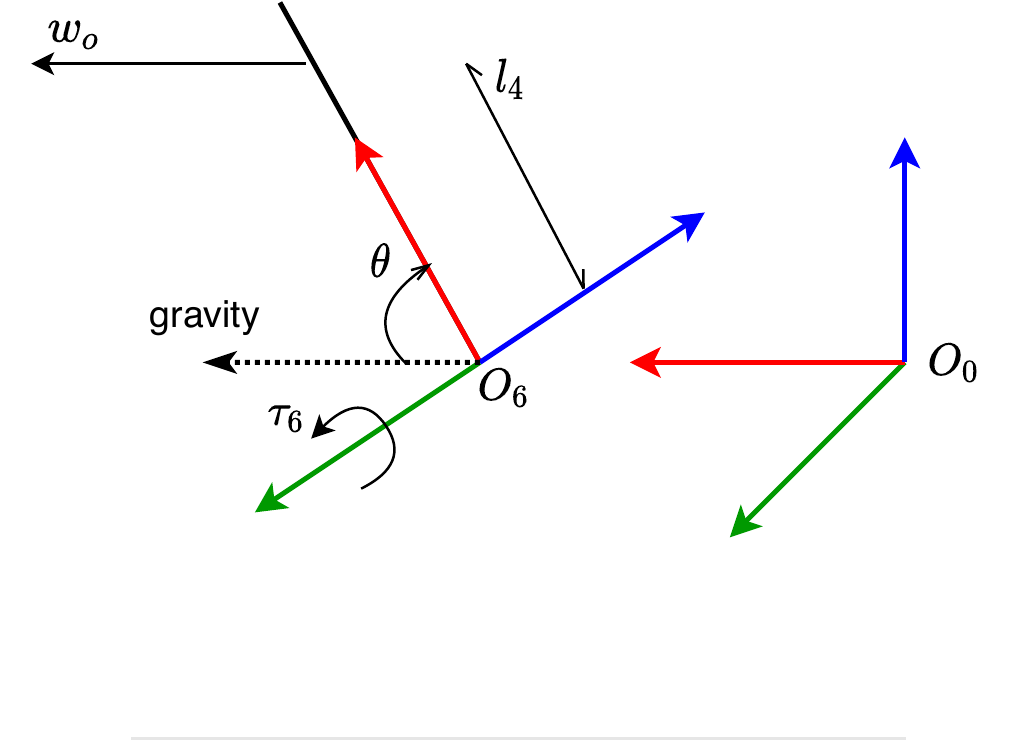}
    \caption{Calculation of $m_o$ from $\tau_6$.}
    \label{fig:rpy}
  \end{subfigure}
  \caption{Variables and frames to calculate $m_o$.}
\end{figure}

Figure \ref{fig:taus} shows torques $\tau_o$ on the different joints. Torques are zero where the weight force is applied along the axis of rotation. In figure \ref{fig:taus} this is the case for joints $O_3$, $O_5$ and $O_7$.

Torques $\tau_o$ are caused by the weight of the object being carried and thus they can be obtained in a form similar to term $G(q)$ in model (\ref{eq:lagrangian}). The mass $m_o$ can be estimated from any component of $\tau_o$, nevertheless, due to the kinematic configuration, it is much easier if we use torque on joint $O_6$.

Consider the figure \ref{fig:rpy}. The frame $O_6$, attached to joint 6, is the frame $O_0$ translated and rotated. To represent the rotation we used the Roll-Pitch-Yaw angles $(\rho, \theta, \psi)$. As it can be seen, the weight $w_o=m_o g$ causes the torque $\tau_6=-m_o g l_4 \sin\theta$ and the mass of the manipulated object can thus be calculated as:
\begin{equation}
  m_o = -\frac{\tau_6}{gl_4\sin\theta}
  \label{eq:mass}
\end{equation}
where $\theta$ is the \textit{pitch} angle of frame $O_6$ w.r.t. $O_0$, $g$ is the gravity acceleration, $\tau_6$ is the disturbance torque calculated according to (\ref{eq:taus}) and $l_4$ is the distance from joint $O_6$ to the center of mass of the manipulated object. As it will be latter explained, errors in the estimation of $l_4$ will cause proportional errors in the estimation of $m_o$.

Note that when $\theta=0$ it is not possible to estimate $m_o$ because $\theta=0$ means that the manipulated object is ``hanging'' from the joint and its weight is not exerting any torque on the joint. There are two possible ways to overcome this situation: use another disturbance $\tau_i$ or moving the manipulator to a useful position. The second option is more feasible since the relation between $w_o$ and $\tau_i$ becomes more complex in the rest of the joints. 

\subsection{Extended Kalman Filter}
Sliding Mode Observer have the great advantage of being robust against disturbances, nevertheless, due to the discontinuos ouput error injection, they can have high frequency components in the estimated states. SMOs are useful for fault reconstruction but, due to the noise and low frequency sampling rate, it is better to use an Extended Kalman Filter (EKF) as part of the closed-loop control strategy.

From (\ref{eq:model1})-(\ref{eq:model2}), we have the state transition model for the noisy system:
\begin{eqnarray}
    \dot{x}_1 &=& x_2 + \nu_1\label{eq:noisymodel1}\\
    \dot{x}_2 &=& f(x_1, x_2, u) + \phi(x_1, x_2, u) + \nu_2\label{eq:noisymodel2}
\end{eqnarray}
where $\nu \in\mathbb{R}^14$ is a vector of noise signals without temporal correlation, zero mean and covariance matrix $Q\in\mathbb{R}^{14\times 14}$. Remember that system state variables are the seven joint positions and seven joint speeds.

In this work, since we are measuring the joint positions, our observation model is:
\begin{equation}
  z = h(x,u) + \omega = [q_1,\dots,q_7]^T + \omega
  \label{eq:obsmodel}
\end{equation}
where $\omega\in\mathbb{R}^7$ is Gaussian noise without temporal correlation, zero mean and covariance matrix $R\in\mathbb{R}^{7\times 7}$.

Chosing between the continuous or discrete version of the EKF depends on the frequency sampling. In this work we achieved a sampling of 250 Hz, which is much faster than the system dynamics, but too slow for a SMO. Also, since Simulink provide tools for continuos controls, we chose the continuos version of the EKF:
\begin{eqnarray}
  \dot{\hat{x}}_1 &=& \hat{x}_2 + K_1 y_1 \label{eq:ekf1}\\
  \dot{\hat{x}}_2 &=& f(\hat{x}_1, \hat{x}_2, u) + K_2 y_2\label{eq:ekf2}\\
  \dot{P} &=& FP + PF^T - KRK^T + Q\label{eq:ekf3}
\end{eqnarray}

where $y=z - \hat{x}_1$ is the error between the estimated and measured outputs, $F$ is the Jacobian of the state transition model and $K$ is the Kalman Gain calculated as:
\[
K = PH^TR^{-1}
\]
with $H\in\mathbb{R}^{7\times 14}$, the Jacobian of the observation model, which, in this case, is the constant matrix:
\[
H = \left[0\quad I_7\right]
\]
Note that in (\ref{eq:ekf2}) we are using only the nominal part of the system and thus, if $\phi\neq 0$ (i.e., if an object is being manipulated), the estimated states will not converge to the real states. Since estimated states are used for position control, such estimation error will cause a steady state error in the controller. This is in principle undesirable, nevertheless, we can tolerate this error since the main goal of this work is the estimation of the mass, not the performance of the controller. Calculations to estimate $m_o$ are made under the assumption that the manipulator is in a constant position, thus, driving the manipulator to a constant $q_f$, although slightly different from the desired $q_g$, is a good enough performance for the control-observation loop.

\subsection{Position control}
As stated before, equations for estimating $m_o$ are derived under the assumption that the manipulator is in a constant configuration. We implemented a PD+Gravity controller using the EKF estimated states:

\begin{equation}
  u = G(\hat{q}) + K_p(q_d - \hat{q}) + K_d(\dot{q}_d - \hat{\dot{q}})
\end{equation}

where $\hat{q}$ and $\hat{\dot{q}}$ are the EKF estimated positions and speeds, respectively; $G(\hat{q})$ is the vector of gravity torques of the nominal part of the model (no noise and no fault signals) and 

As explained in the previous section, when the manipulator takes an object, the estimated states do not converge to the real ones and then the controller shows an steady state error. Nevertheless, this is tolerable since the goal of the proposed system is not the control performance but the observer performance for the mass estimation.

To improve control performance, a trajectory is planned using a 5th degree polynomial. This type of trajectory allows to set initial and final position, speed and acceleration, which is ideal for the manipulation tasks. It is important to note that since the trajectories are calculated in joint space, the cartesian movement of the end effector doesn not result in a straight line. Although algorithms to plan movement task for manipulators are out of the scope of this research, we used this approach to simplify the tuning of the control constants for experimentation.

\section{Simulink implementation}
\label{sec:implementation}
\begin{figure}[h!]
  \centering
  \includegraphics[width=0.9\textwidth]{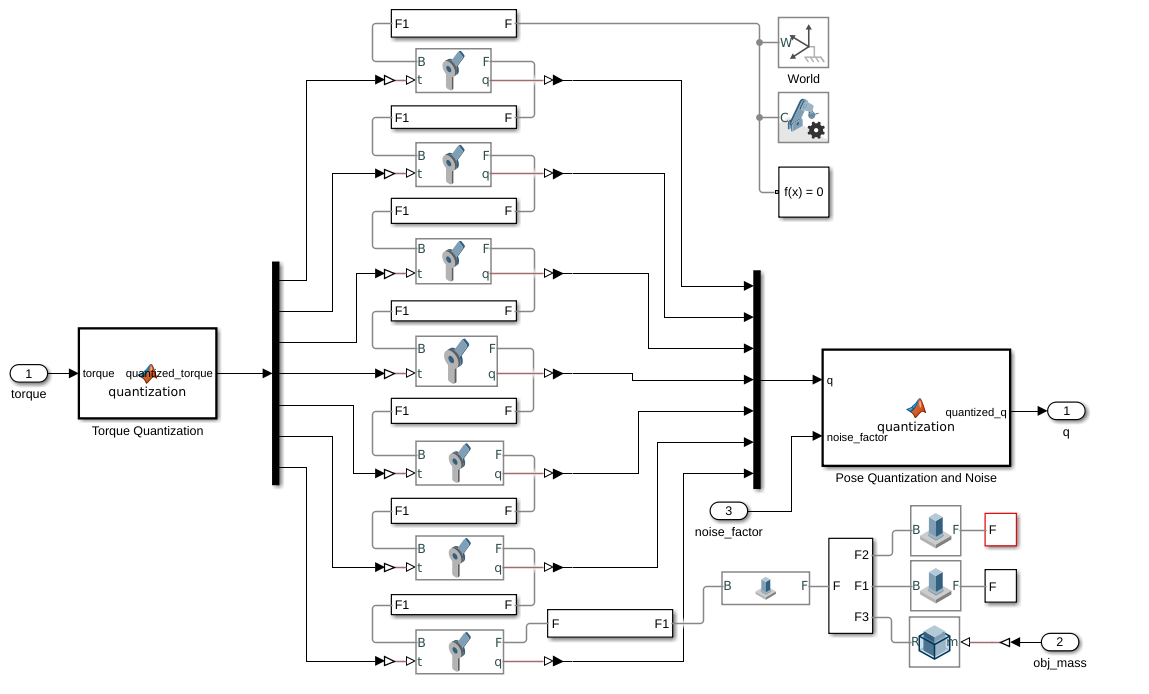}
  \caption{The resulting model after importing URDF}\
  \label{fig:SimulinkModel}
\end{figure}

Implementation of control and observer was made in Simulink mainly using the Robotics Systems Toolbox, Simscape and ROS Toolbox. We followed the suggested steps in \cite{matlab_designing, matlab_trajectory}. To reproduce this proposal in other robots, we recommend to read first such tutorials. In the following subsection we roughly describe this implementation.

\subsubsection*{Simulator}
Most of the tests were performed first in simulation. To such purpose, we used the \texttt{smimport} function of Simscape Multibody to generate the corresponding rigid body tree. Figure \ref{fig:SimulinkModel} shows the resulting block diagram. As it can be seen, simulation of robot dynamics is separated from control and observation. Communication between manipulator and controller and observer is made via ROS topics in order to keep them transparent to the hardware. This way, switching between real and simulated manipulator is easy.

\subsubsection*{Quantization and noise}
\begin{figure}[h!]
  \centering
  \begin{subfigure}[b]{0.4\textwidth}
    \centering
    \includegraphics[width=\textwidth]{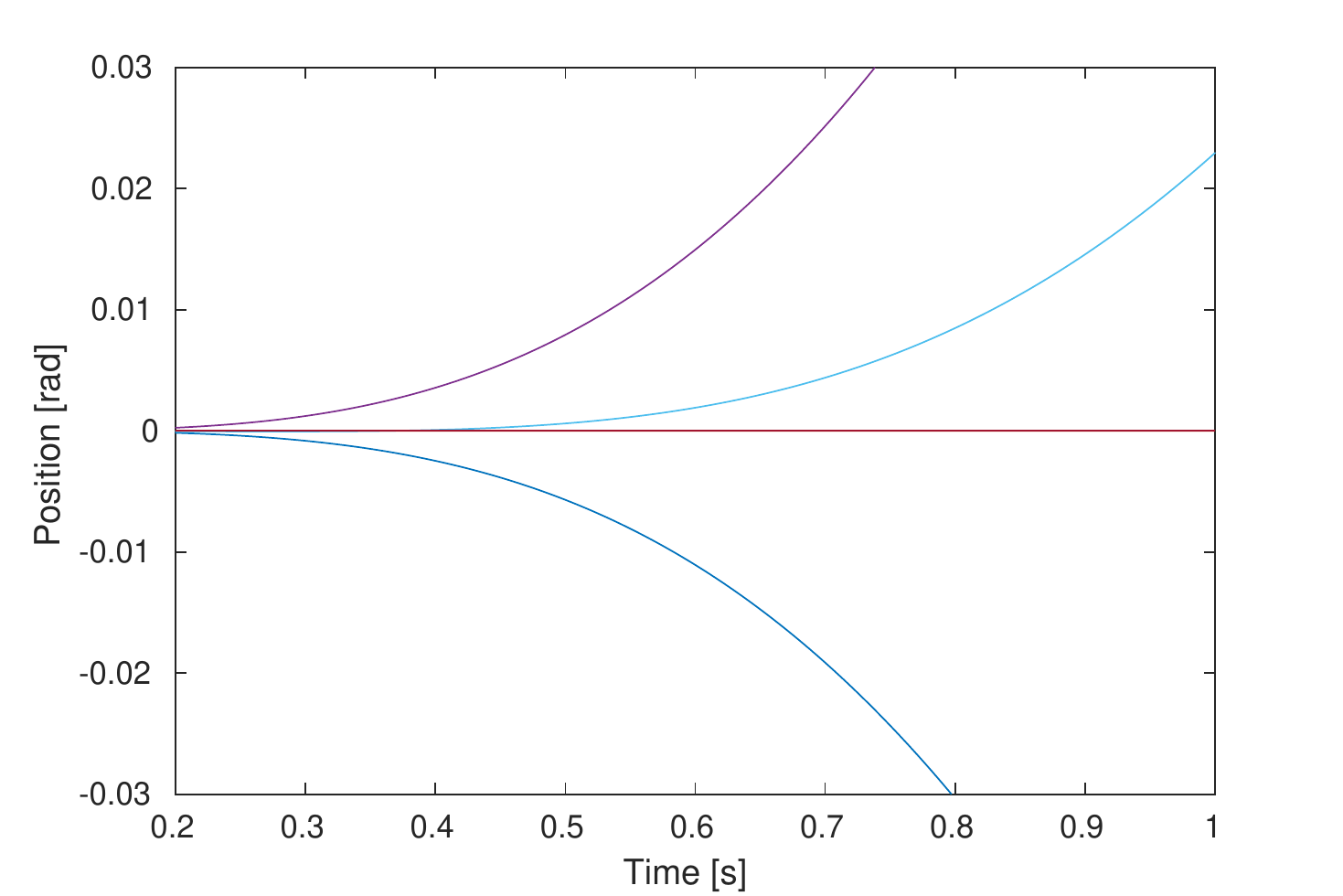}
    \caption{Continuos joint positions}
    \label{fig:q_quantized1}
  \end{subfigure}
  \begin{subfigure}[b]{0.4\textwidth}
    \centering
    \includegraphics[width=\textwidth]{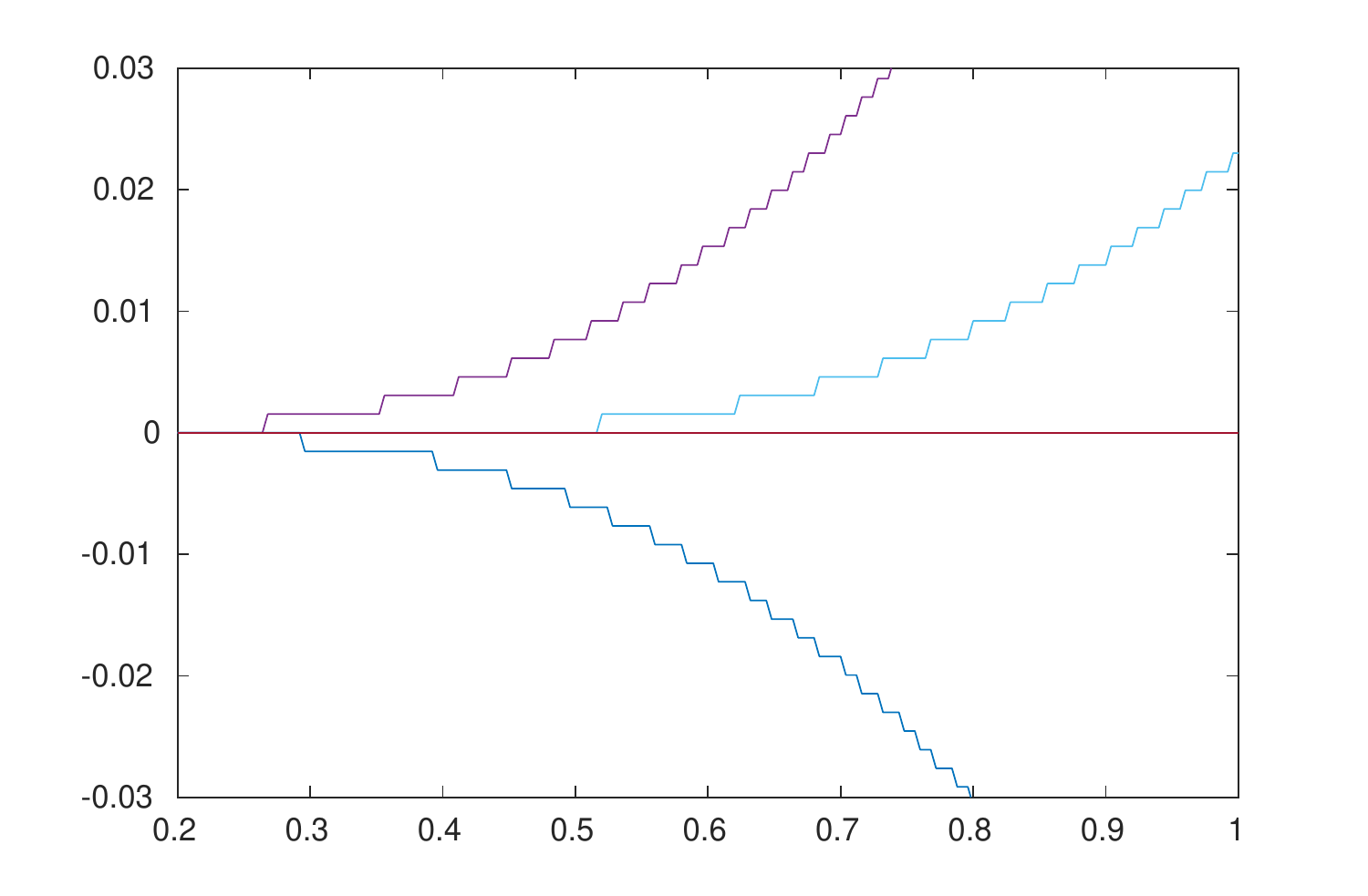}
    \caption{Quantized joint position}
    \label{fig:q_quantized2}
  \end{subfigure}
  \begin{subfigure}[b]{0.4\textwidth}
    \centering
    \includegraphics[width=\textwidth]{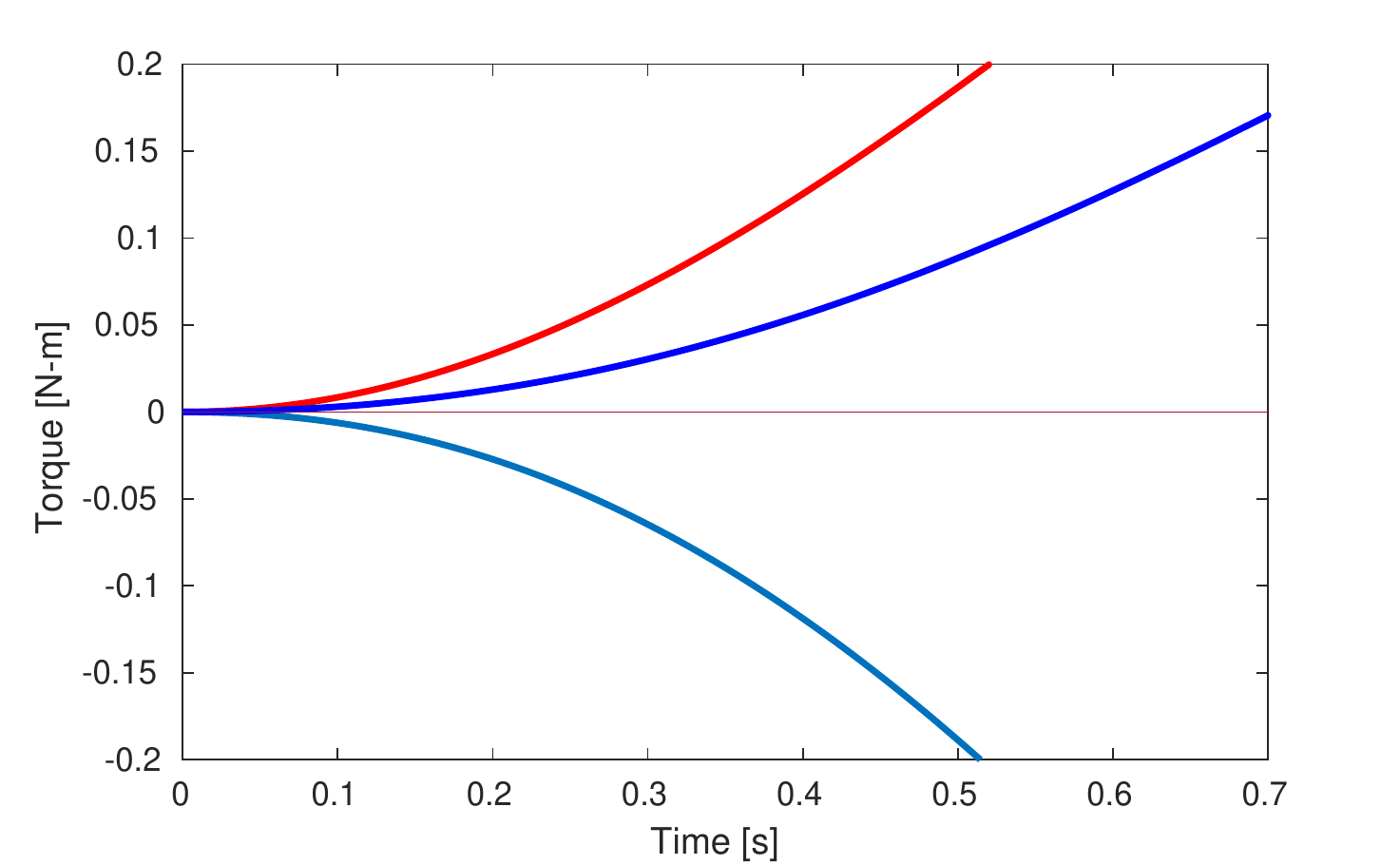}
    \caption{Continuos input torque}
    \label{fig:torque_quantized1}
  \end{subfigure}
  \begin{subfigure}[b]{0.4\textwidth}
    \centering
    \includegraphics[width=\textwidth]{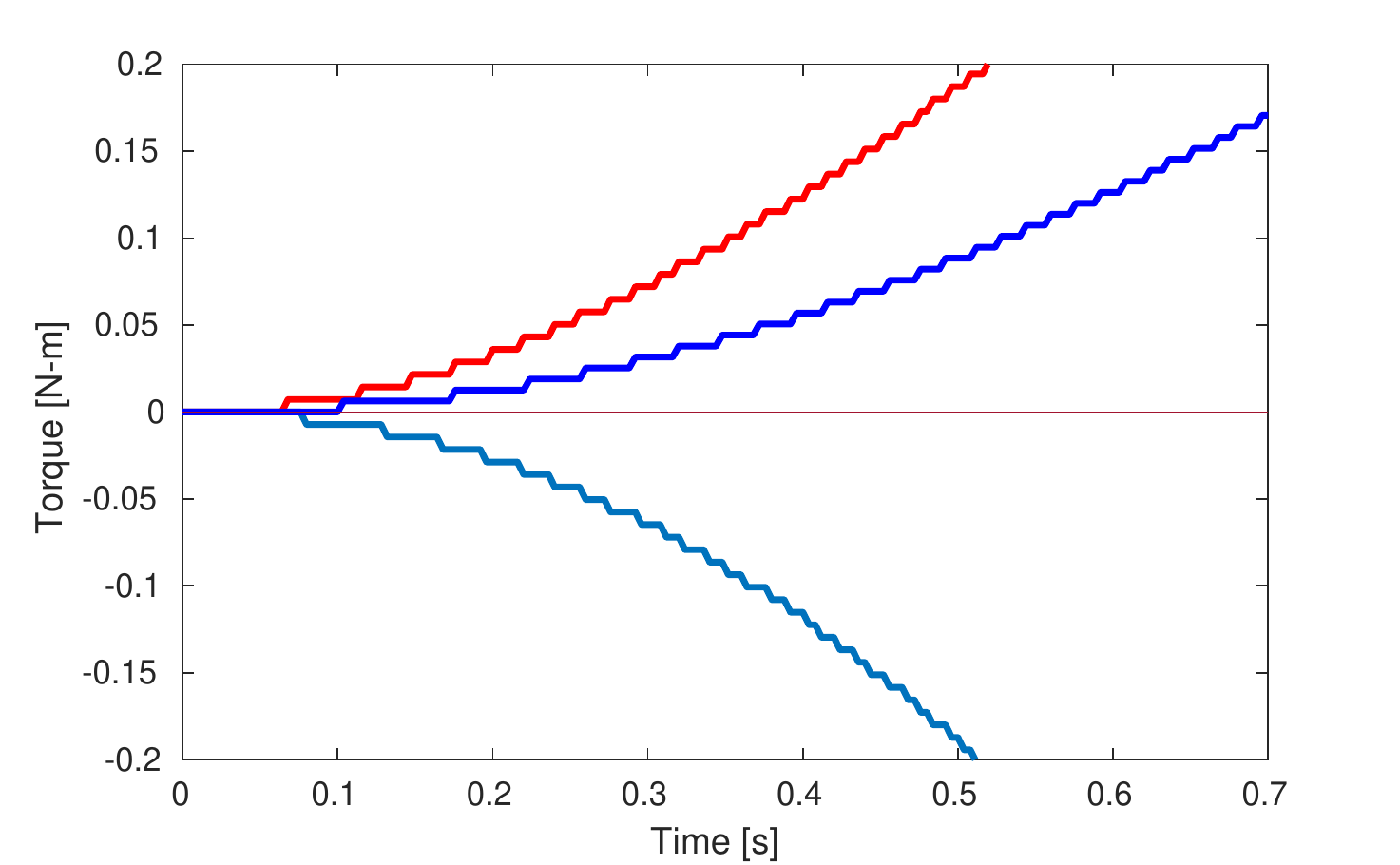}
    \caption{Quantized input torque}
    \label{fig:torque_quantized2}
  \end{subfigure}
  \caption{Position and torque quantization}
  \label{fig:quantization}
\end{figure}

To achieve a simulation more similar to the conditions of the real manipulator, we added a quantization for input torque and position measurements. Dynamixel motors can read position with 12 bit resolution for 360 degrees and can set torque with 10 bit resolution. Torque resolution is not constant since it depends on motor model and battery level. The implemented simulation adds this quantization effect. An example of such quantization is shown in figure \ref{fig:quantization}. Also, noise was added to joint measurements. We added a noise equivalent to $\pm 2$ bits. 

\subsubsection*{Observation and control}
\begin{figure}
  \centering
  \begin{subfigure}{0.45\textwidth}
  \centering
  \includegraphics[width=\textwidth]{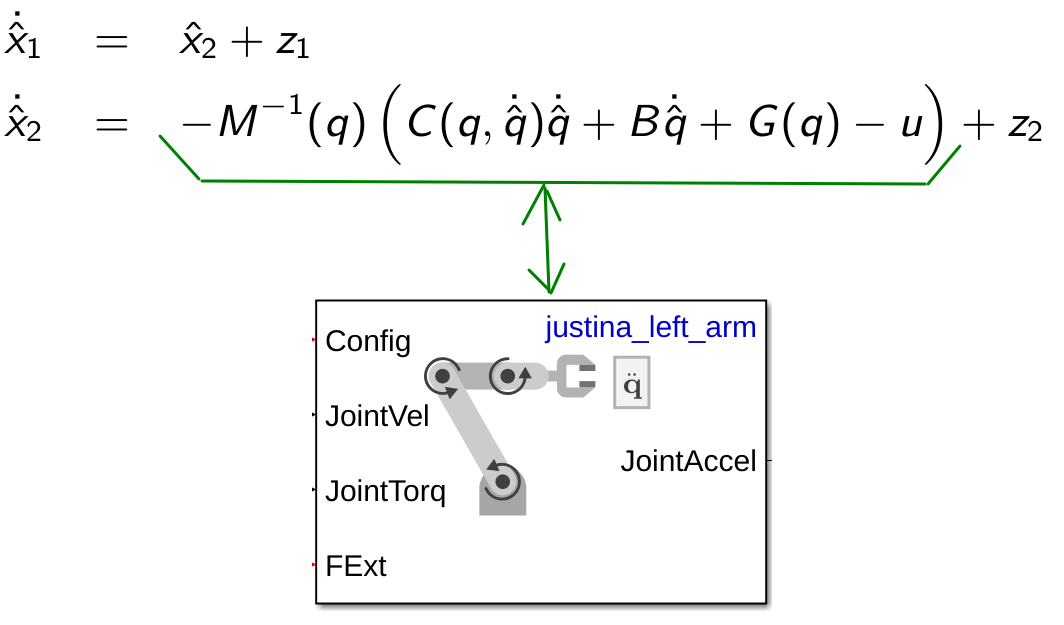}
  \caption{Forward Dynamics block for SMO}
  \label{fig:SimulinkSMO}
  \end{subfigure}
  \hfill
  \begin{subfigure}{0.48\textwidth}
  \centering
  \includegraphics[width=0.5\textwidth]{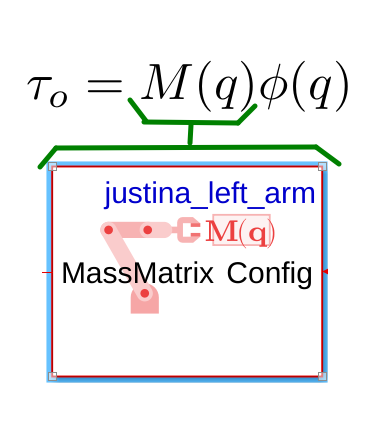}
  \caption{Mass Matrix block in mass estimation}
  \label{fig:SimulinkMassEst}
  \end{subfigure}
  \begin{subfigure}{0.45\textwidth}
  \centering
  \includegraphics[width=\textwidth]{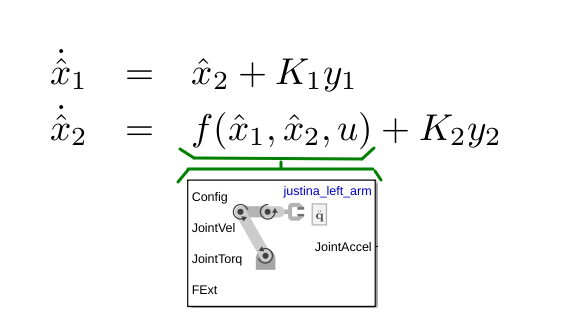}
  \caption{Forward Dynamics block for EKF}
  \label{fig:EKFSimulink}
  \end{subfigure}
  \hfill
  \begin{subfigure}{0.45\textwidth}
  \centering
  \includegraphics[width=\textwidth]{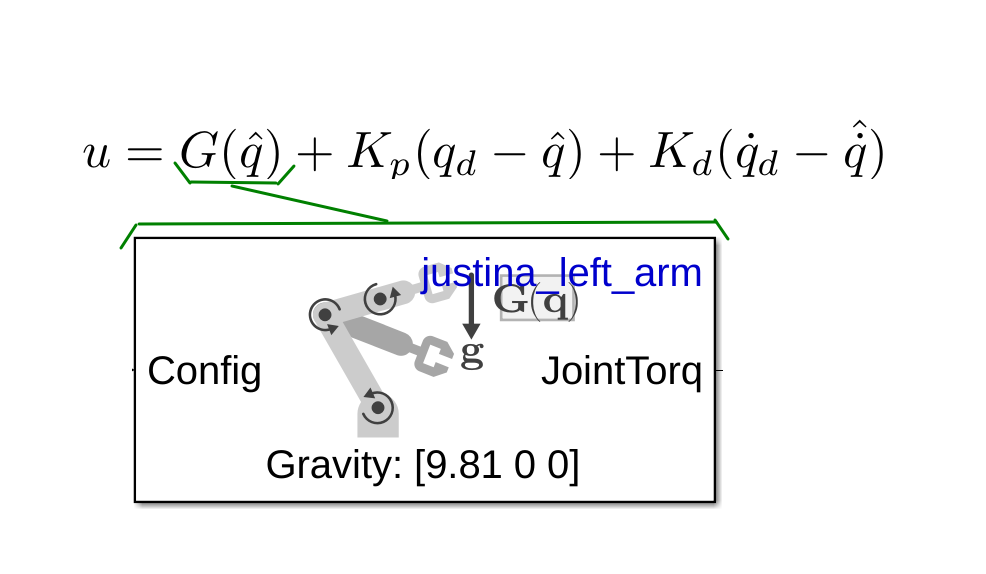}
  \caption{Gravity block for PD+Control}
  \label{fig:ControlSimulink}
  \end{subfigure}
  \caption{Implementation using Simulink Blocks}
\end{figure}

The Sliding Mode Observer is a copy of the system plus an output error injection term, with the form:
\begin{eqnarray}
  \dot{\hat{x}}_1 &=& \hat{x}_2 + z_1\label{eq:observer1}\\
  \dot{\hat{x}}_2 &=& -M^{-1}(q)\left(C(q, \dot{\hat{q}})\dot{\hat{q}} + B\dot{\hat{q}} + G(q) - u\right) + z_2\label{eq:observer2}
\end{eqnarray}
  The nominal part can be derived from the lagrangian, nevertheless, the analytic form is too complex due to the number of DOF. Instead, we used the numeric solution provided by the Inverse Dynamics block from the Simulink Robotics System Toolbox library, as shown in figure \ref{fig:SimulinkSMO}. Also, from equation (\ref{eq:taus}), it is necessary to compute the inertia matrix $M(q)$ to obtain the perturbing torques $\phi$. Similar to the observer, instead of obtaining the algebraic expression of $M(q)$, we used the numeric calculation provided by the Robotics System Toolbox, as shown in figure \ref{fig:SimulinkMassEst}.

\begin{figure}[h!]
  \centering
  \includegraphics[width=\textwidth]{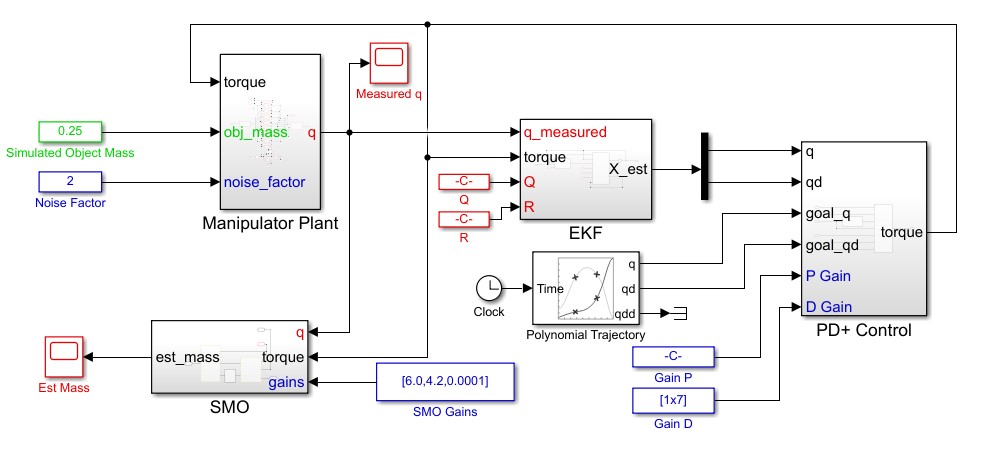}
  \caption{Block diagram of the full system implemented in Simulink}
  \label{fig:FullSystemBlocks}
\end{figure}

Similar to the SMO, the EKF is a copy of the system plus an output error injection term, but in this case, such term is not discontinuous, but it is a linear gain of the error, with such linear gain calculated to minimize noise effect. As shown in figure \ref{fig:EKFSimulink}, the forward dynamics term in equations (\ref{eq:ekf1})-(\ref{eq:ekf2}) is calculated using the corresponding Robotics System Toolbox block. Figure \ref{fig:FullSystemBlocks} shows the block diagram of the full system as implement in Simulink. 

\subsubsection*{A ROS node for each task}
\begin{figure}[h!]
  \centering
  \includegraphics[width=0.95\textwidth]{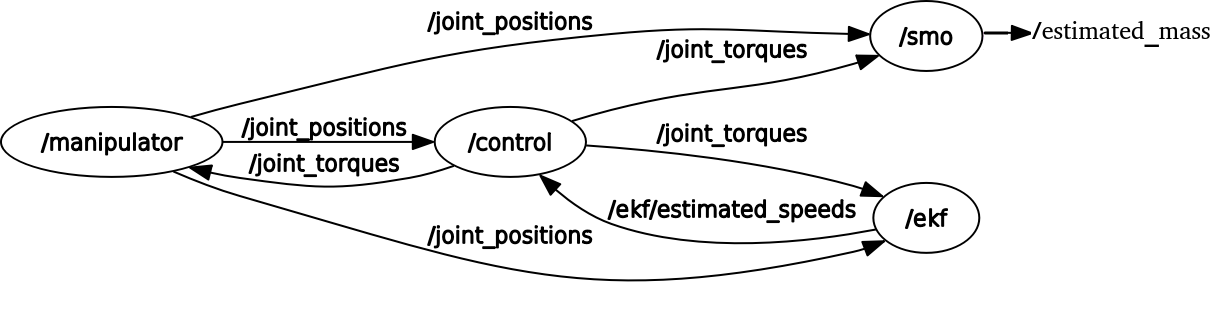}
  \caption{Graph diagram of ROS-nodes communications}
  \label{fig:RosNodes}
\end{figure}

To ease implementation with the real manipulator and integration with the rest of the subsystems, we separated every task in different ROS nodes: arm dynamics simulation, control, SMO and EKF. Signals are shared via topics and ROS parameters are used for all tuning constants. Figure \ref{fig:RosNodes} shows the connections between the different nodes and the ROS topics they communicate through. 

\section{Simulation results}
\label{sec:results}
\subsection{Description of the experiment}
\begin{figure}[h!]
  \centering
  \includegraphics[width=0.7\textwidth]{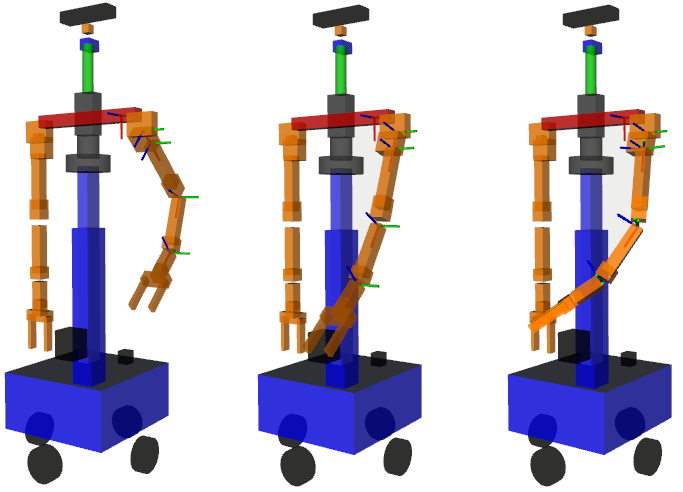}
  \caption{Positions used for testing masss estimation}
  \label{fig:test_poses}
\end{figure}

Mass estimation was tested moving the manipulator to several positions commonly used for grasping objects. Figure \ref{fig:test_poses} shows these predefined positions. It is worth to note that mass estimation will work in any configuration as long as wrist pitch $\theta$ (see equation (\ref{eq:mass})) is different from zero. In a real application, it is better to estimate the mass only when $\vert\theta\vert > \theta_T$ where $\theta_T$ is a threshold large enough to avoid large errors due to the division by a very small value of $\theta$. In this work, we used $\theta_T=0.2$. 

Frequency sampling was set to 250 Hz. This value is the frequency achieved with the real manipulator. Dynamixel servomotors were configured to 1 Mbps of baudrate and we have nine motors in the manipulator: seven motors for the seven degrees of freedom, and two more for the manipulator. Considering the total number of bytes to be sent and received, the required time is in the order of 1 ms, nevertheless, smallest configurable USB latency in Ubuntu is 1 ms, which needs to be added to the time for sending and receiving RS485 data. 4 ms was chosen as approximately twice the time required to send torques and read positions from all motors. This frequency was set both for tests with real and simulated manipulator. 

Experiment for mass estimation was made simulating an object with mass $m=0.25$ located at the center of the gripper. We used the constants shown in table \ref{tab:constants}, where $I_{14}$ and $I_7$ represent identity matrices of orders 14 and 7 respectively. To filter the output error injection term of the SMO we used a 4th degree Butterworth Low-Pass filter with cutoff frequency of 1 Hz. 
\begin{table}[h!]
  \centering
  \begin{tabular}{|c|c|c|c|}
    \hline
    SMO & EKF & Control & Mass Est\\
    \hline
    $\lambda=6.0$ & $Q = 0.003I_{14}$ & $K_p= 2.5$ & $l_4 = 0.21$\\
    $\alpha=4.2$ & $R = 0.001I_7$ & $K_d = 0.5$  & \\
    \hline
  \end{tabular}
  \caption{Constants for control and observation}
  \label{tab:constants}
\end{table}

\subsection{Mass estimation}
\begin{figure}
  \centering
  \includegraphics[width=0.45\textwidth]{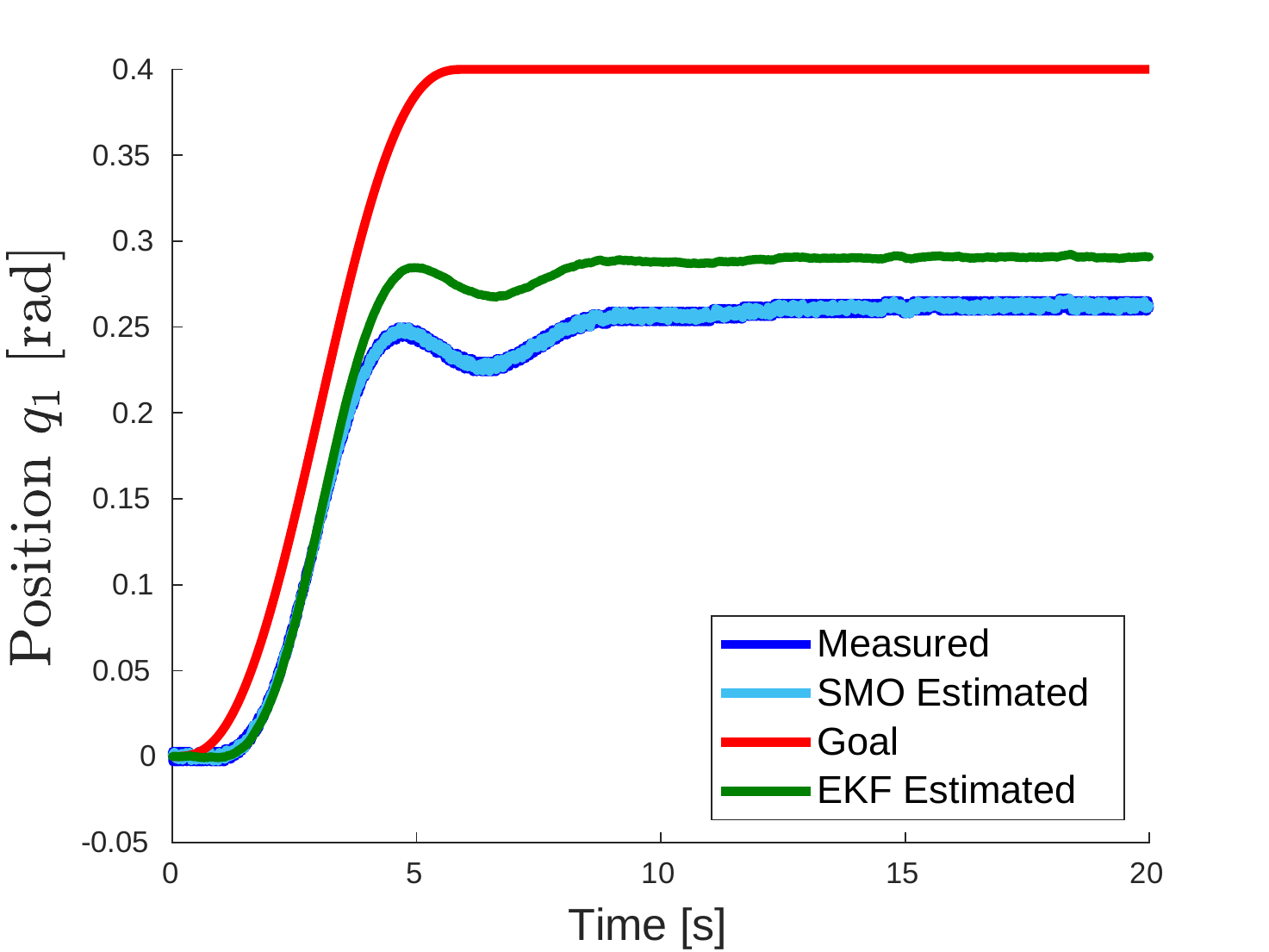}
  \includegraphics[width=0.45\textwidth]{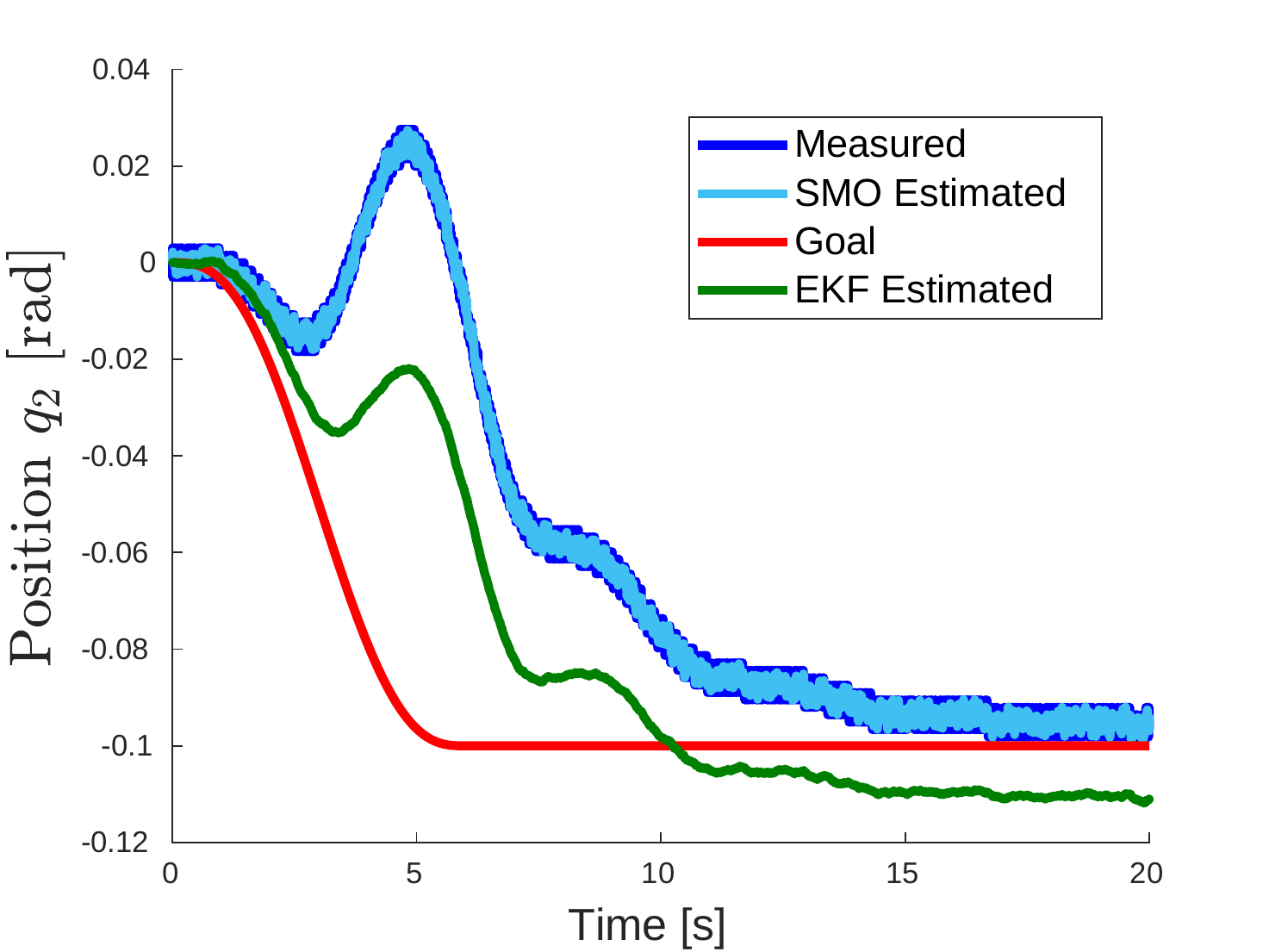}
  \includegraphics[width=0.45\textwidth]{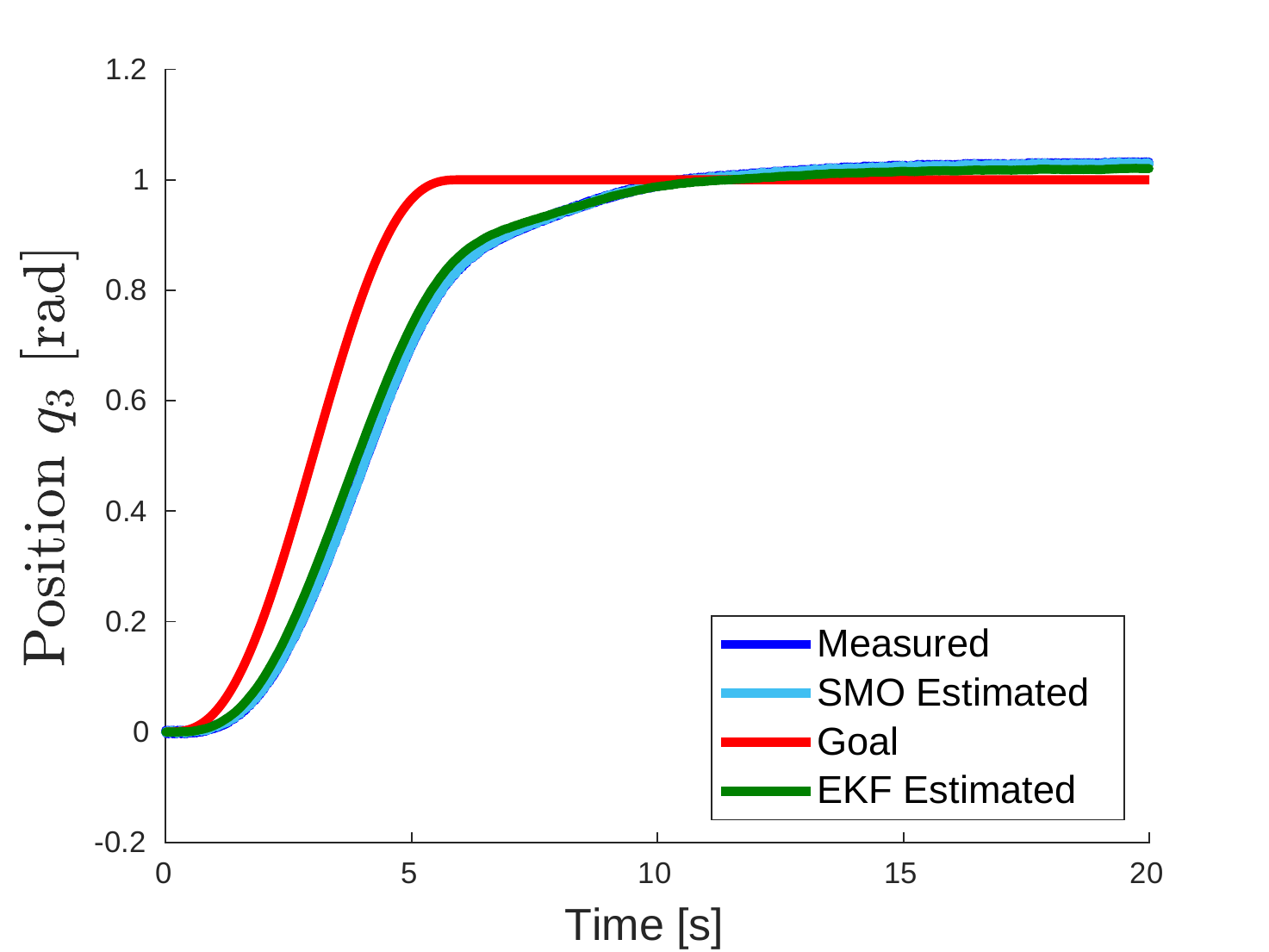}
  \includegraphics[width=0.45\textwidth]{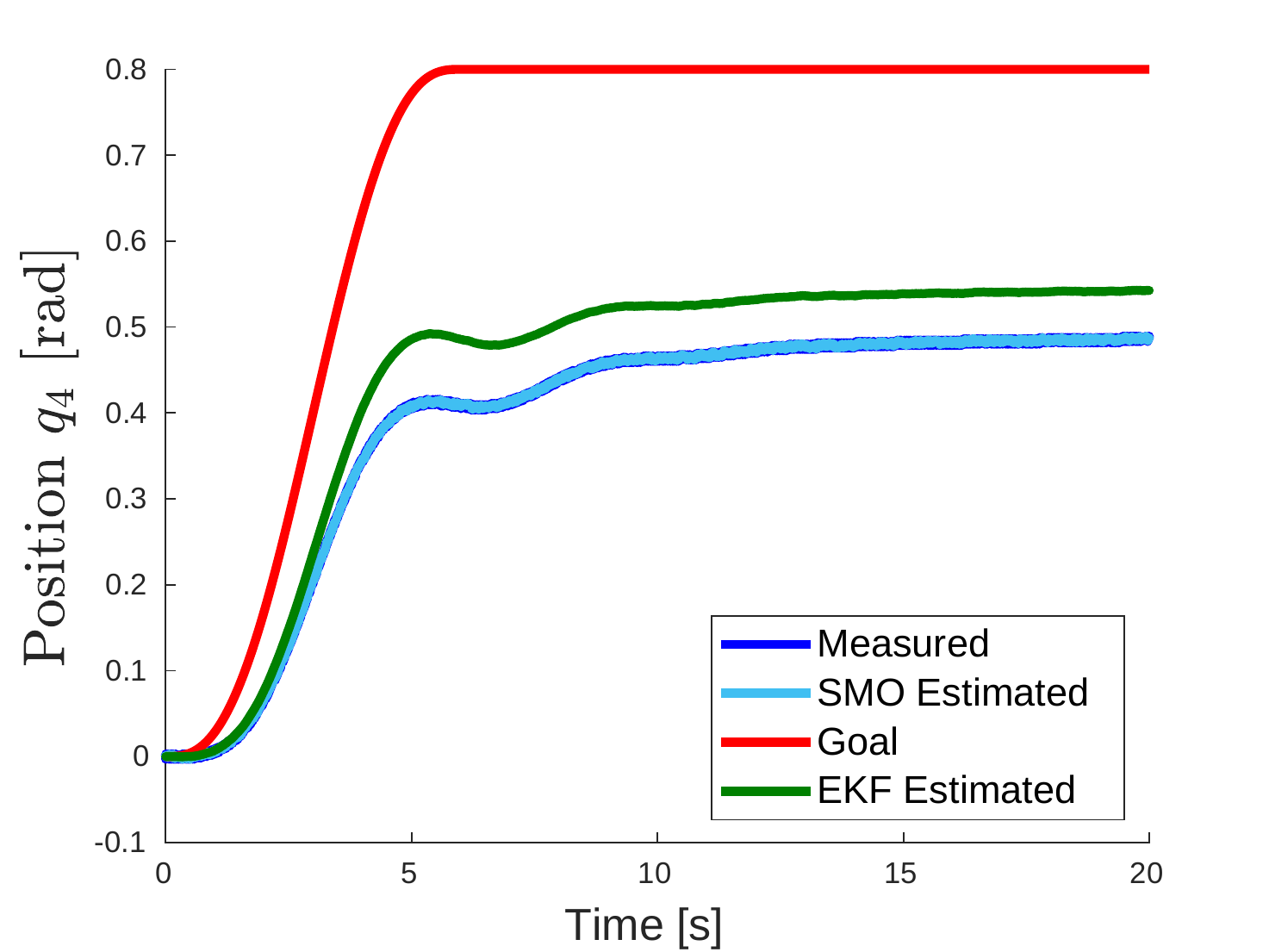}
  \includegraphics[width=0.45\textwidth]{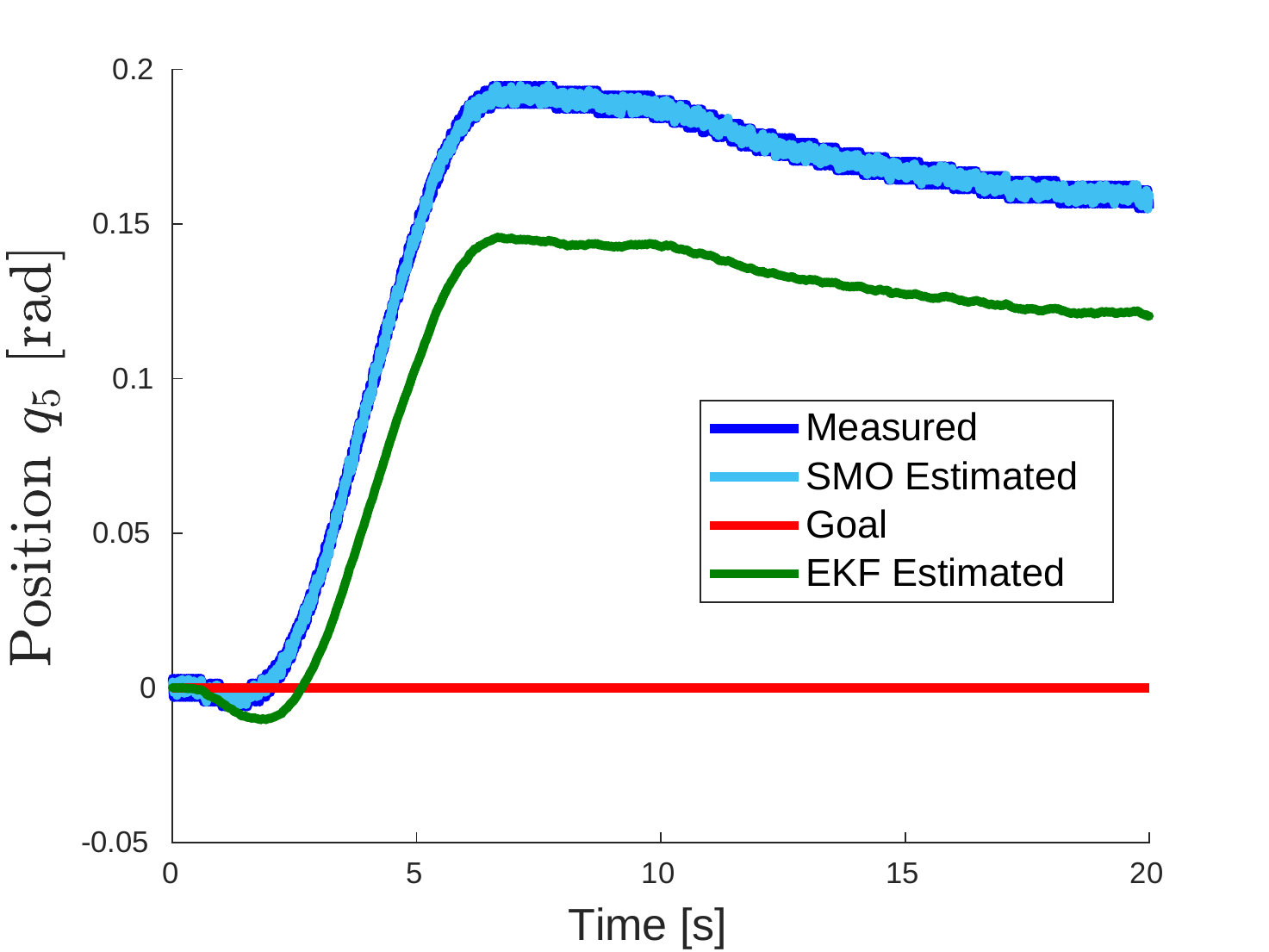}
  \includegraphics[width=0.45\textwidth]{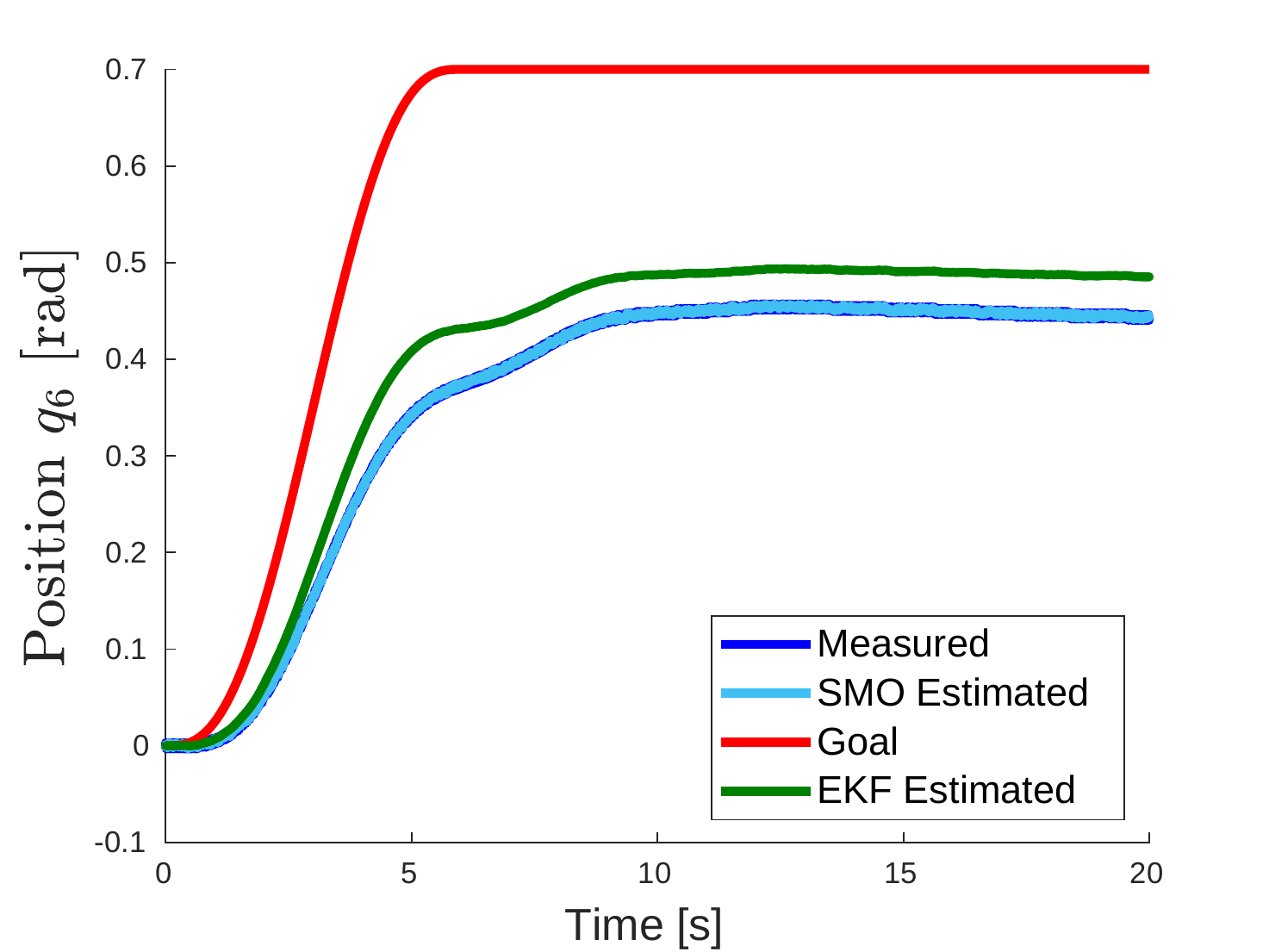}
  \includegraphics[width=0.45\textwidth]{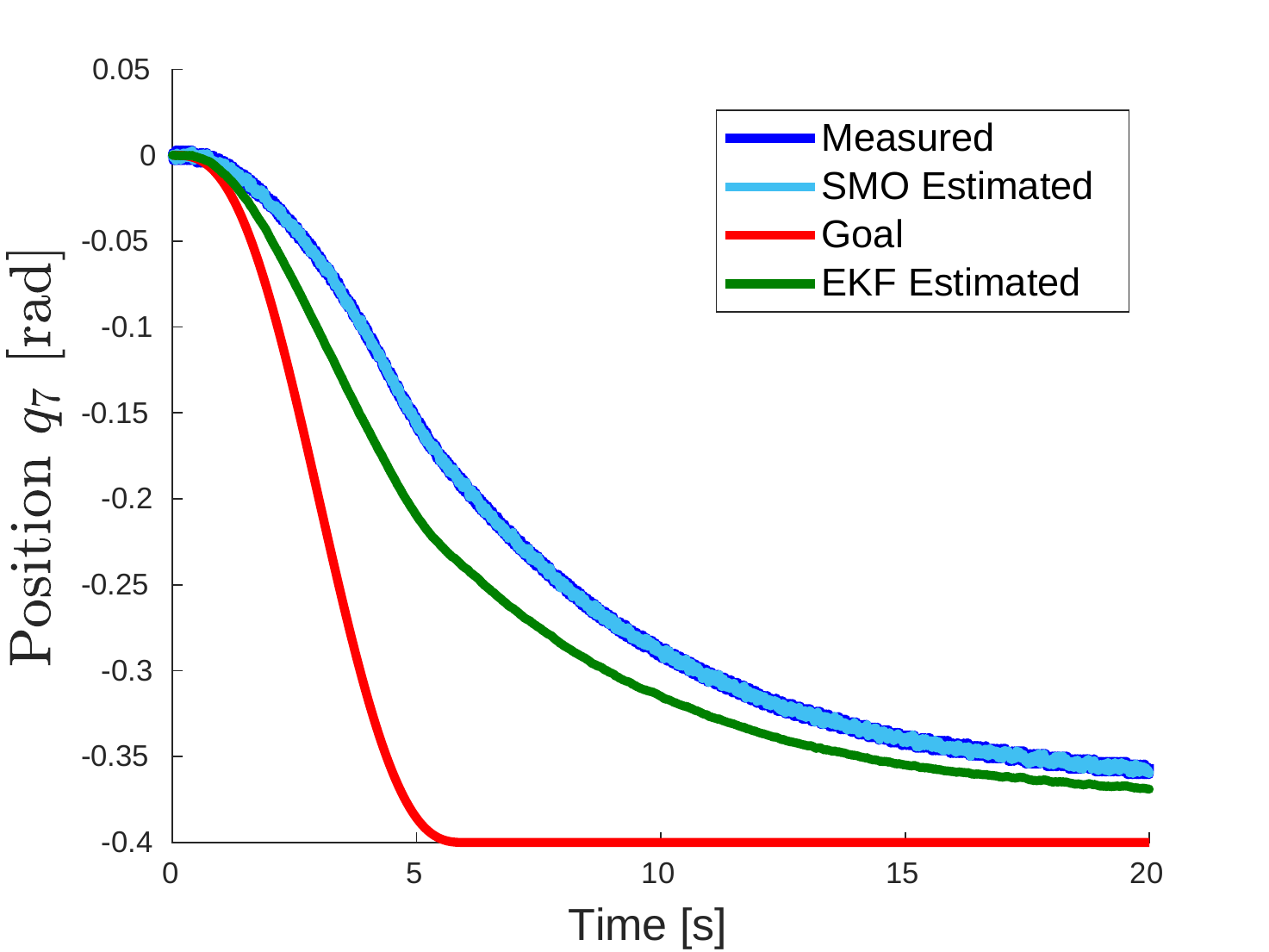}
  \caption{Comparison of measured, goal and estimated positions.}
  \label{fig:result_q}
\end{figure}

Figure \ref{fig:result_q} shows the results for the estimation and filtering of the angular position while moving the arm to one of the predefined positions. As it can be observed, the measured position is noisy and does not converge to the desired position, nevertheless, this is an expected behavior. As explained before, the objective of this work is the mass estimation of the manipulated object, not the performance of the controller. In the presence of the fault signal (an object in the end effector), the controller shows steady state error. Comparing the estimated positions between the SMO and the EKF, we can see that the SMO-estimated positions perfectly track the measured positions, since in general, SMOs are designed to be robust against faults and disturbances, nevertheless, SMO-estimated positions show the problem of chattering. Instead, EKF-estimated positions show an steady-state error but do not show chattering nor noise, making them more suitable for implementing the PD+ control. 

\begin{figure}
  \centering
  \includegraphics[width=0.45\textwidth]{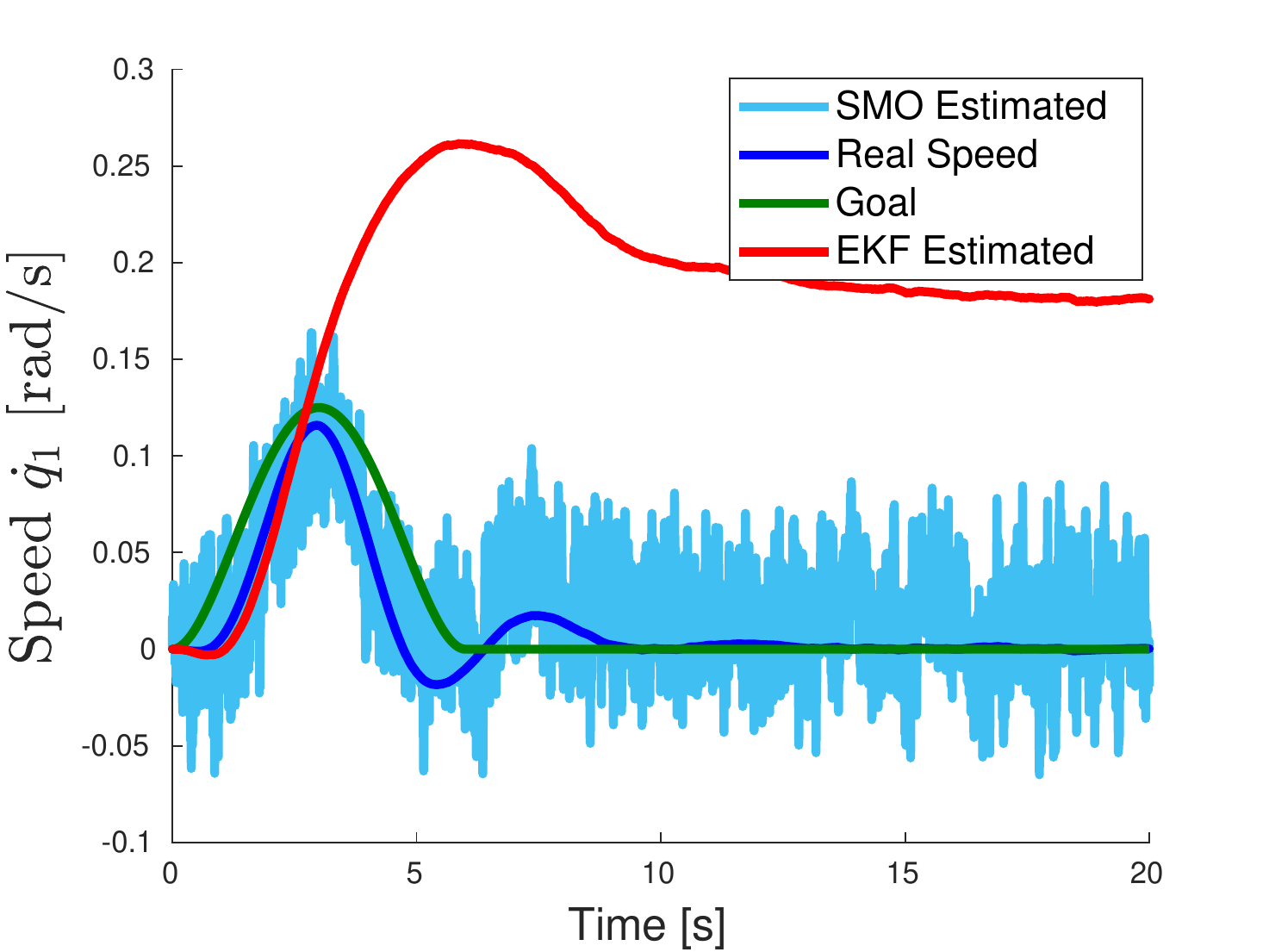}
  \includegraphics[width=0.45\textwidth]{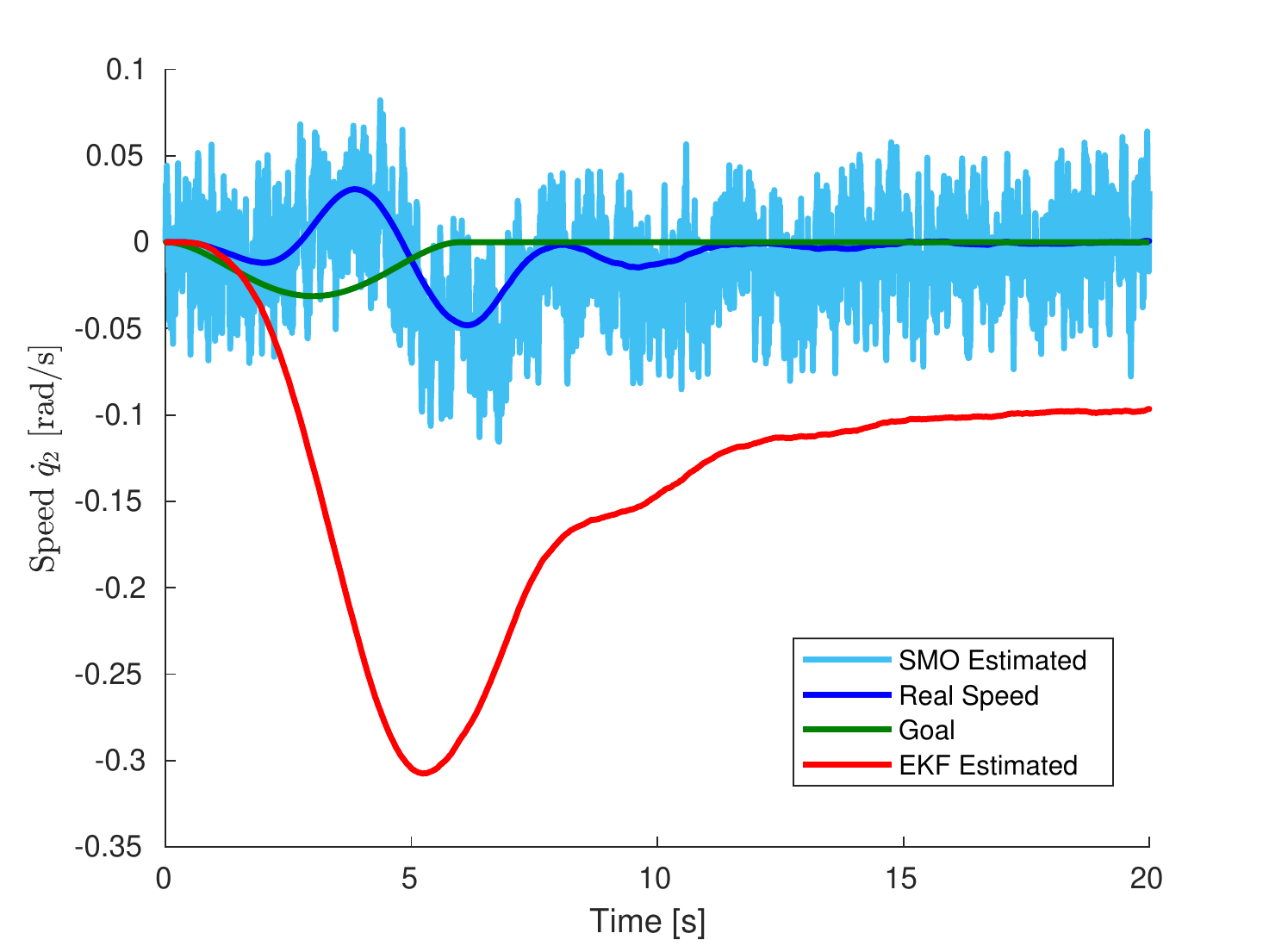}
  \includegraphics[width=0.45\textwidth]{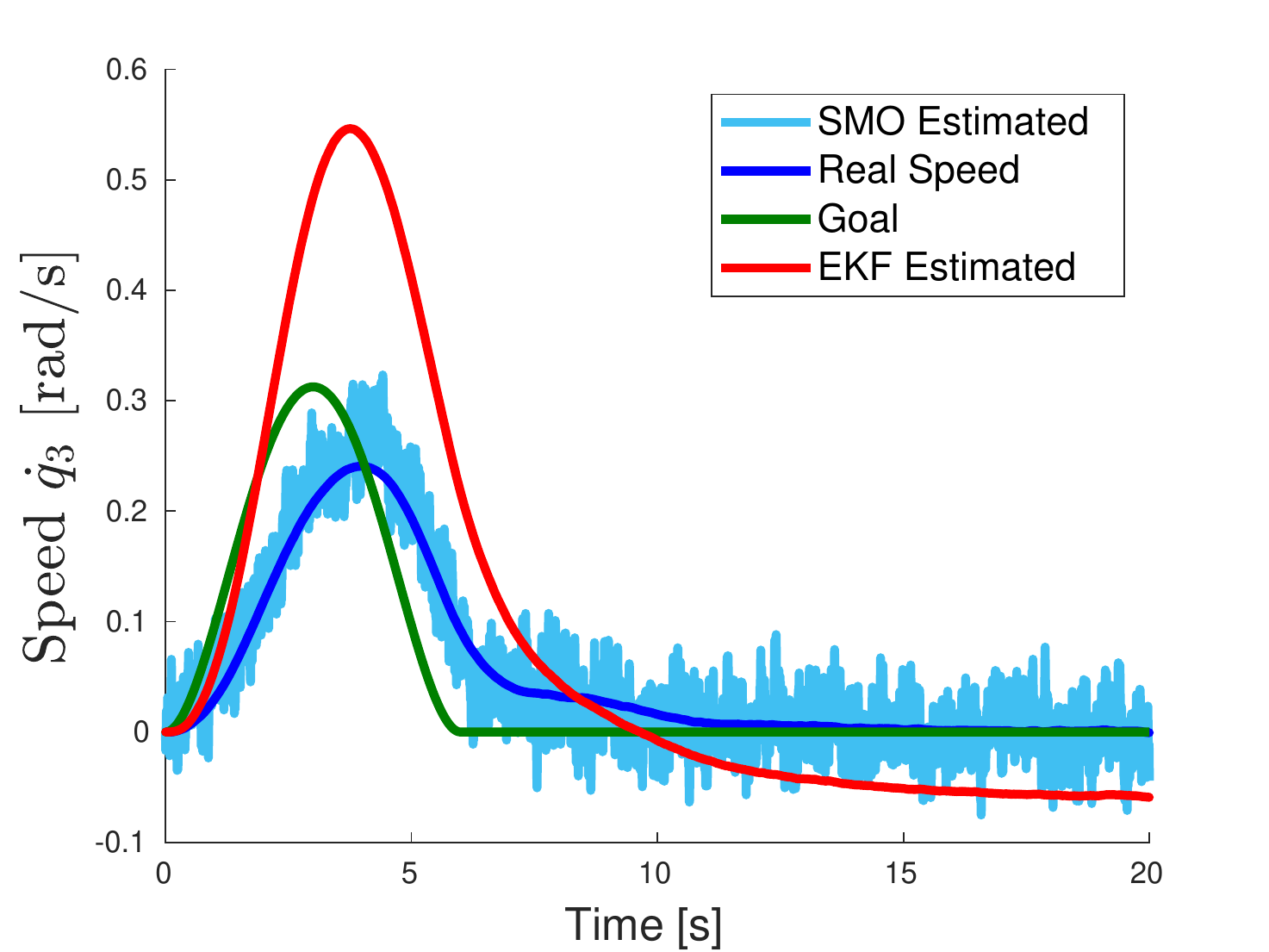}
  \includegraphics[width=0.45\textwidth]{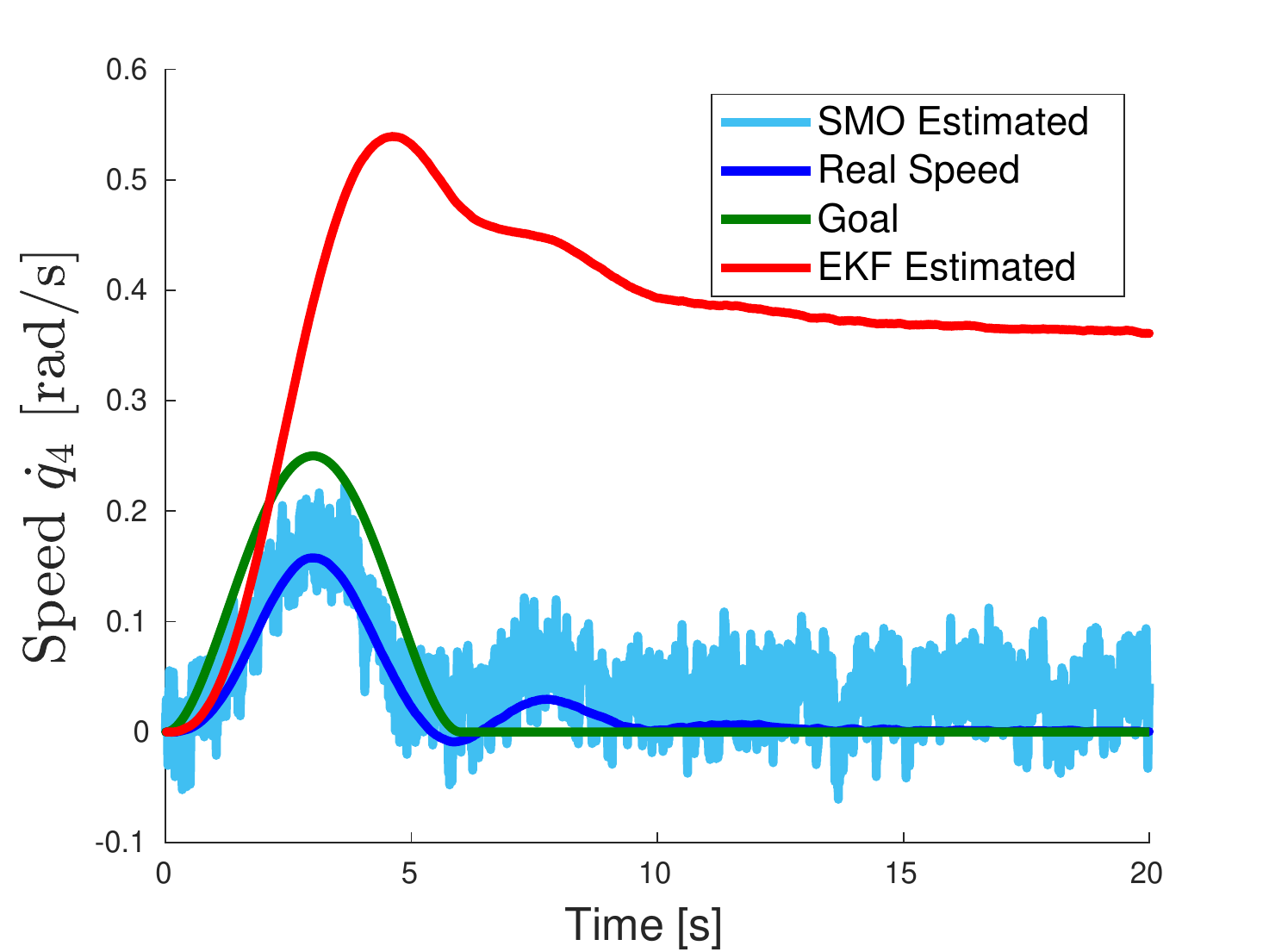}
  \includegraphics[width=0.45\textwidth]{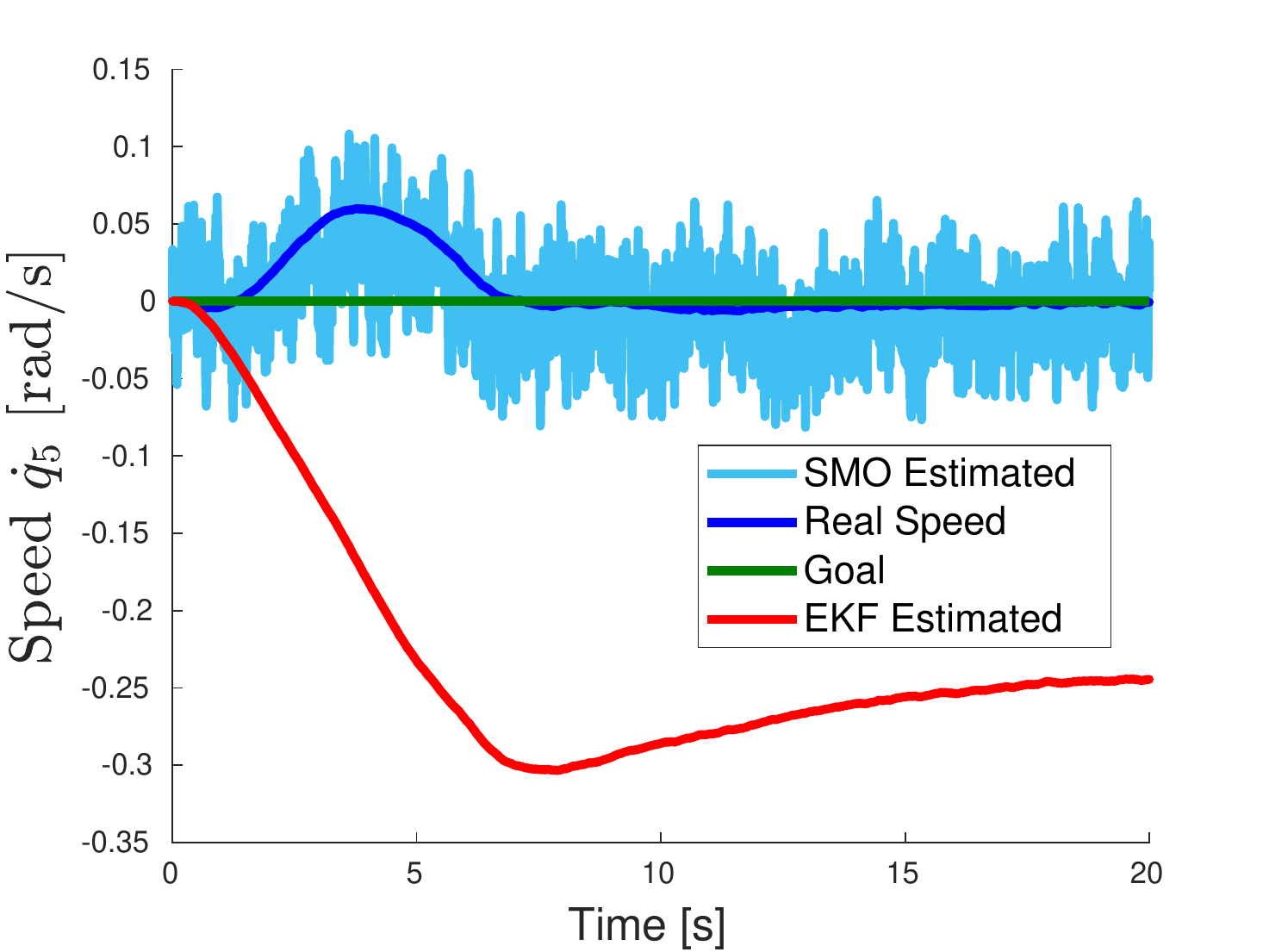}
  \includegraphics[width=0.45\textwidth]{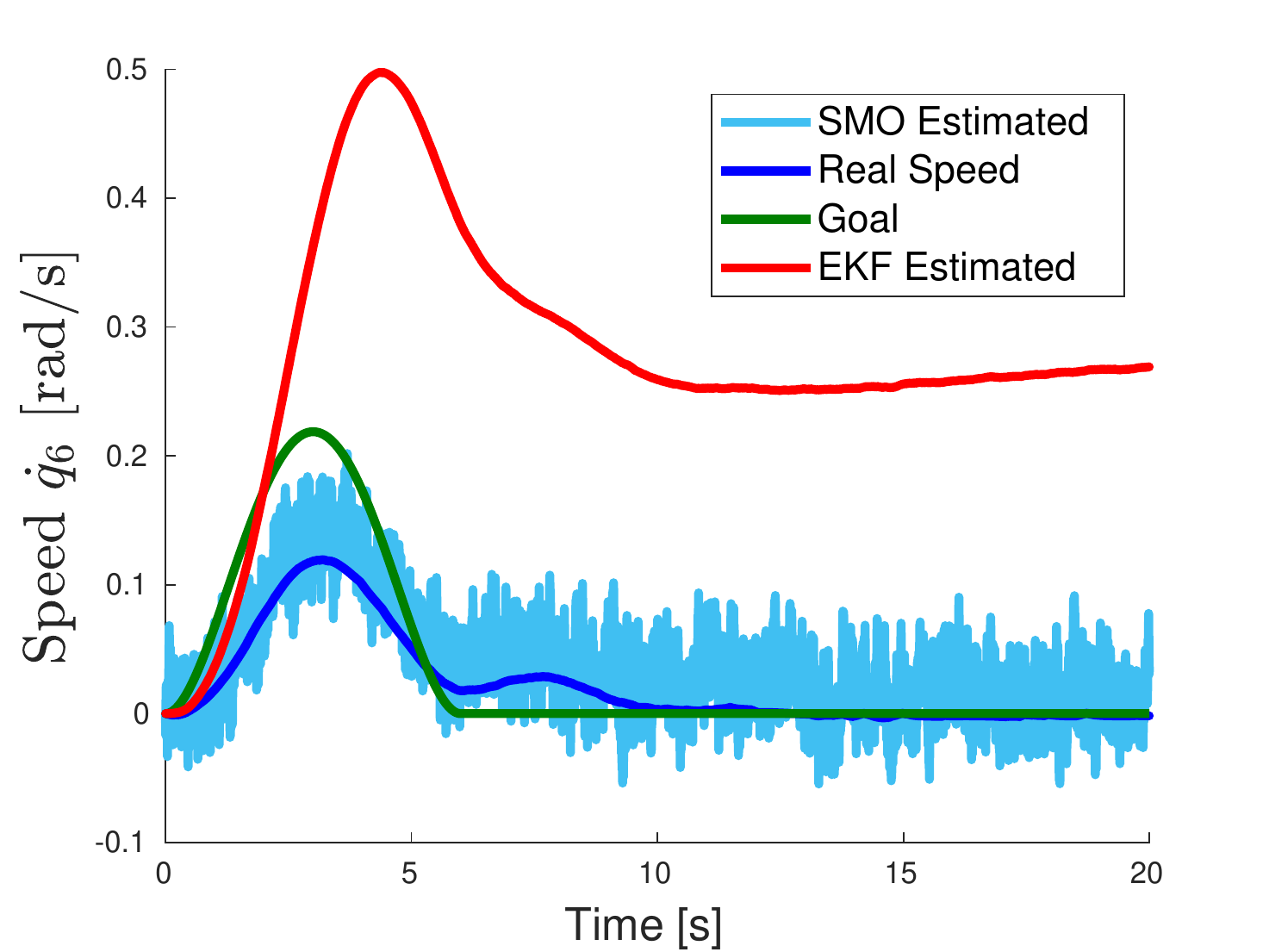}
  \includegraphics[width=0.45\textwidth]{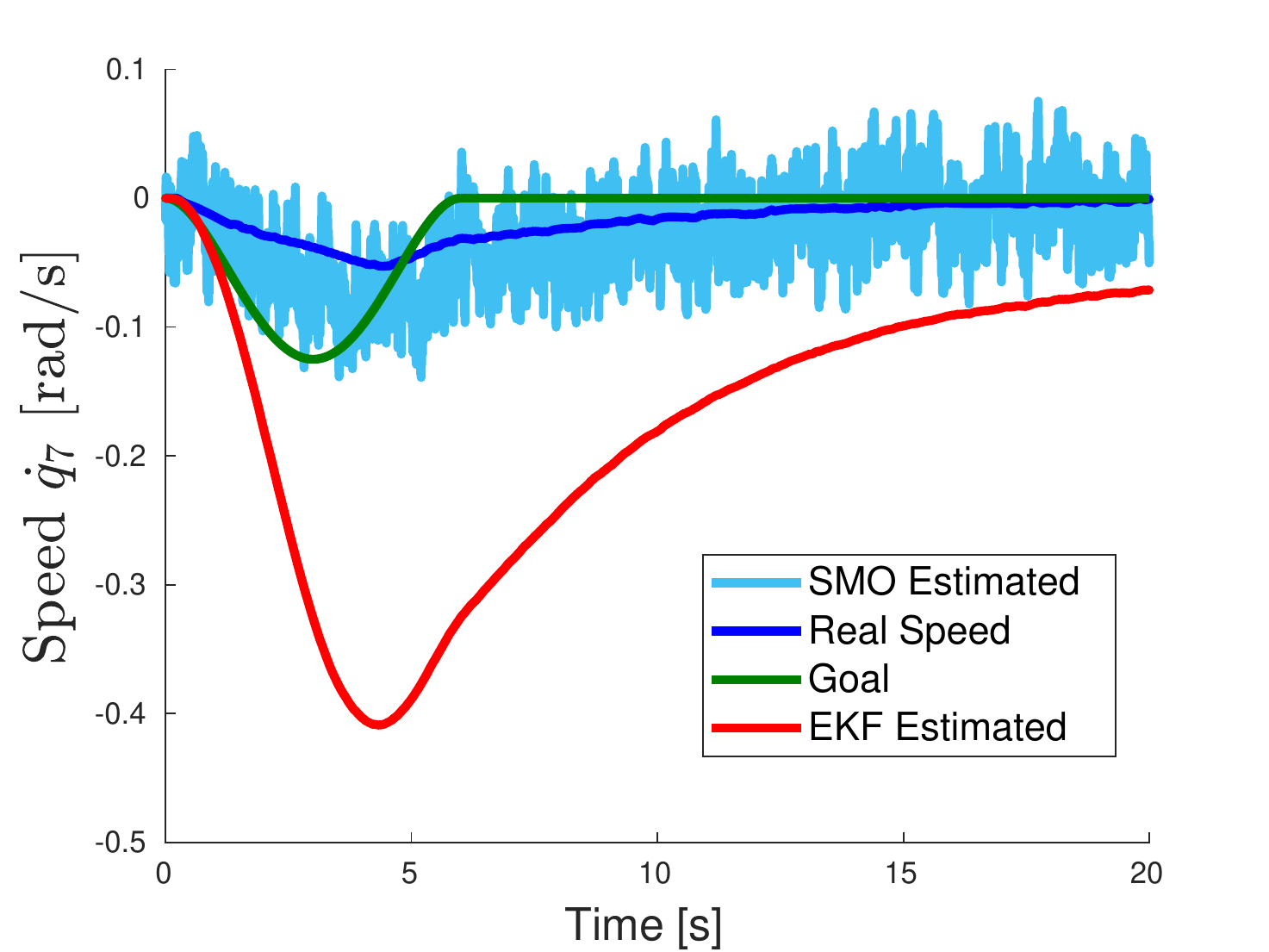}
  \caption{Comparison of simulated, goal and estimated joint speeds.}
  \label{fig:result_qp}
\end{figure}

Figure \ref{fig:result_qp} shows the results for the estimation of the angular speeds. Although in the real manipulator we don't have a measurement of the joint speed, we included the simulated speed for testing purposes. As it can be seen, current speed does not converge to the desired one. Same as the angular position, this error is due to the simplicity of the controller. Similar to the positions, the SMO-estimated speeds are nearer to the current speeds but with the problem of big chattering. Instead, the EKF-estimated speeds show a big error but without noise, which makes them better to be used in the controller. 

\begin{figure}
  \centering
  \includegraphics[width=0.5\textwidth]{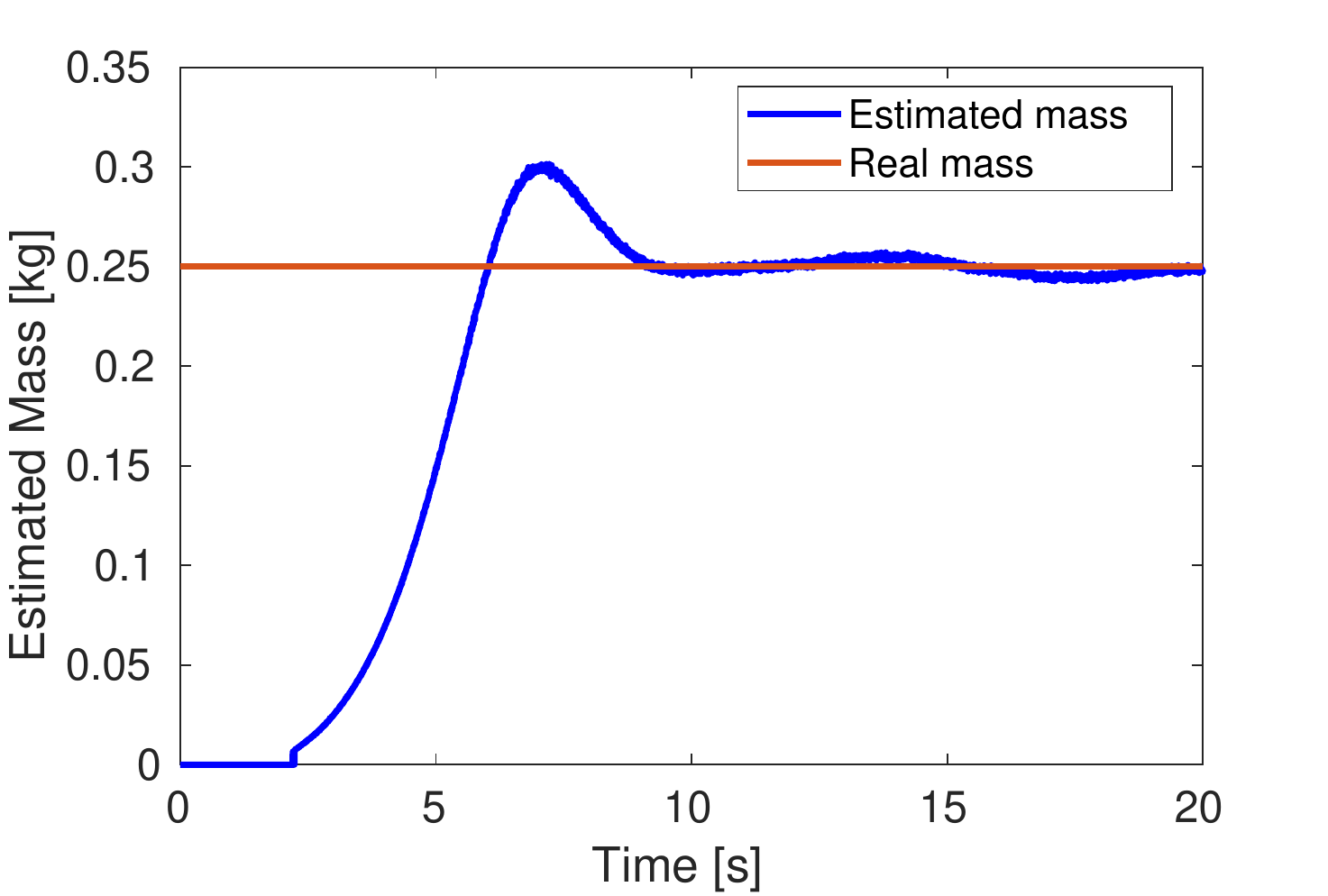}
  \caption{Mass estimation results.}
  \label{fig:MassEstimation}
\end{figure}

Finally, figure \ref{fig:MassEstimation} shows the resulting estimated mass. As explained in section \ref{sec:MassEstimation}, in this work we make the mass estimation only when the manipulator is in a constant position. Ideally, estimations should be made only when $\dot{q}=0$, nevertheless, this is a hard condition. Instead, we set estimated mass to zero if $\Vert\dot{q}\Vert > 0.5$, that is why in figure \ref{fig:MassEstimation} the estimation is equal to zero for the first few seconds. As soon as the arm stops, the estimated value converges to 0.25 kg, the actual mass of the object. 

\subsection{Sources of estimation errors}
\begin{figure}
  \centering
  \begin{subfigure}{0.3\textwidth}
    \centering
    \includegraphics[width=\textwidth]{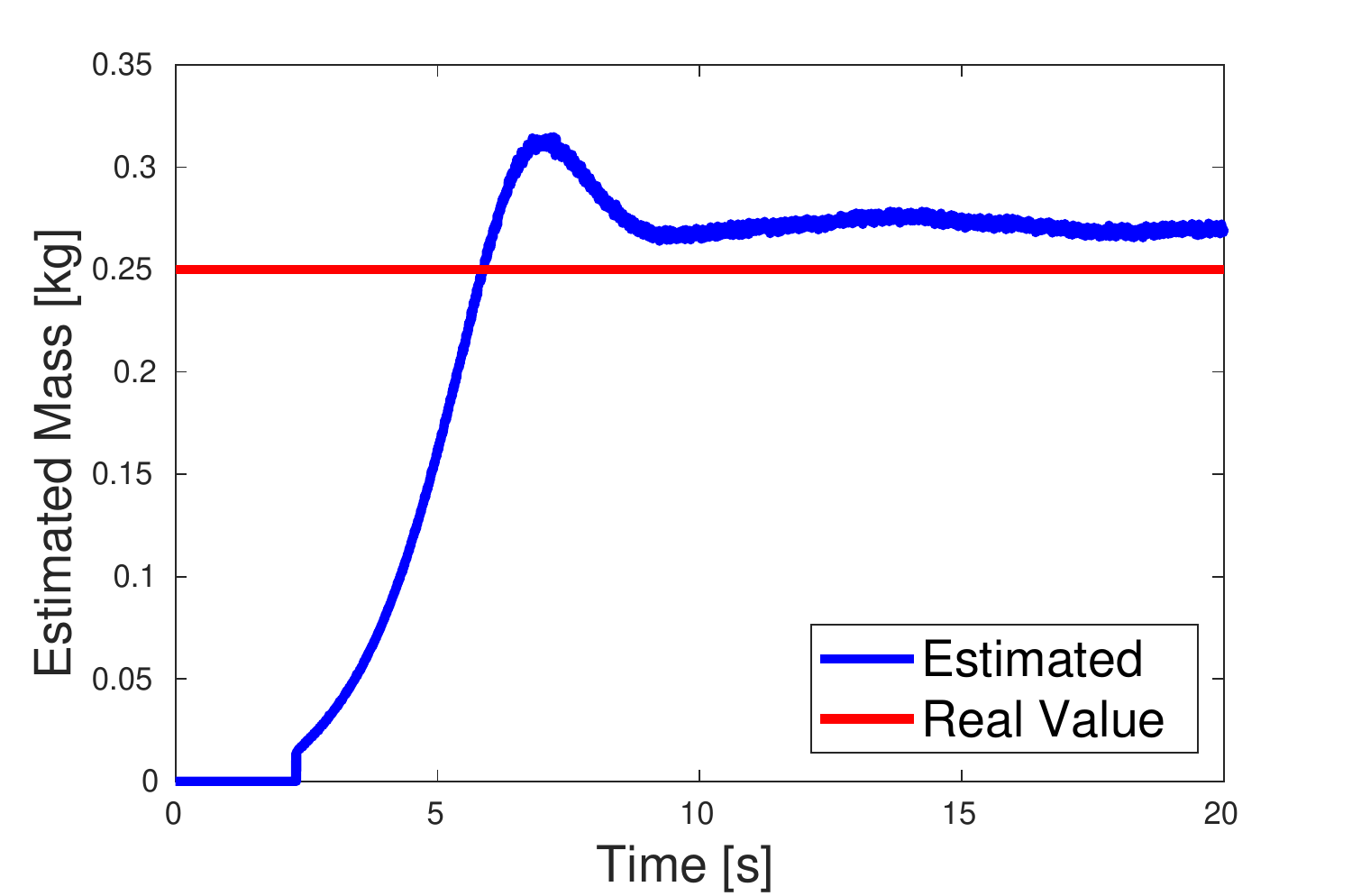}
    \caption{Undervalued 20\%}
    \label{fig:param_error2}
  \end{subfigure}
  \begin{subfigure}{0.3\textwidth}
    \centering
    \includegraphics[width=\textwidth]{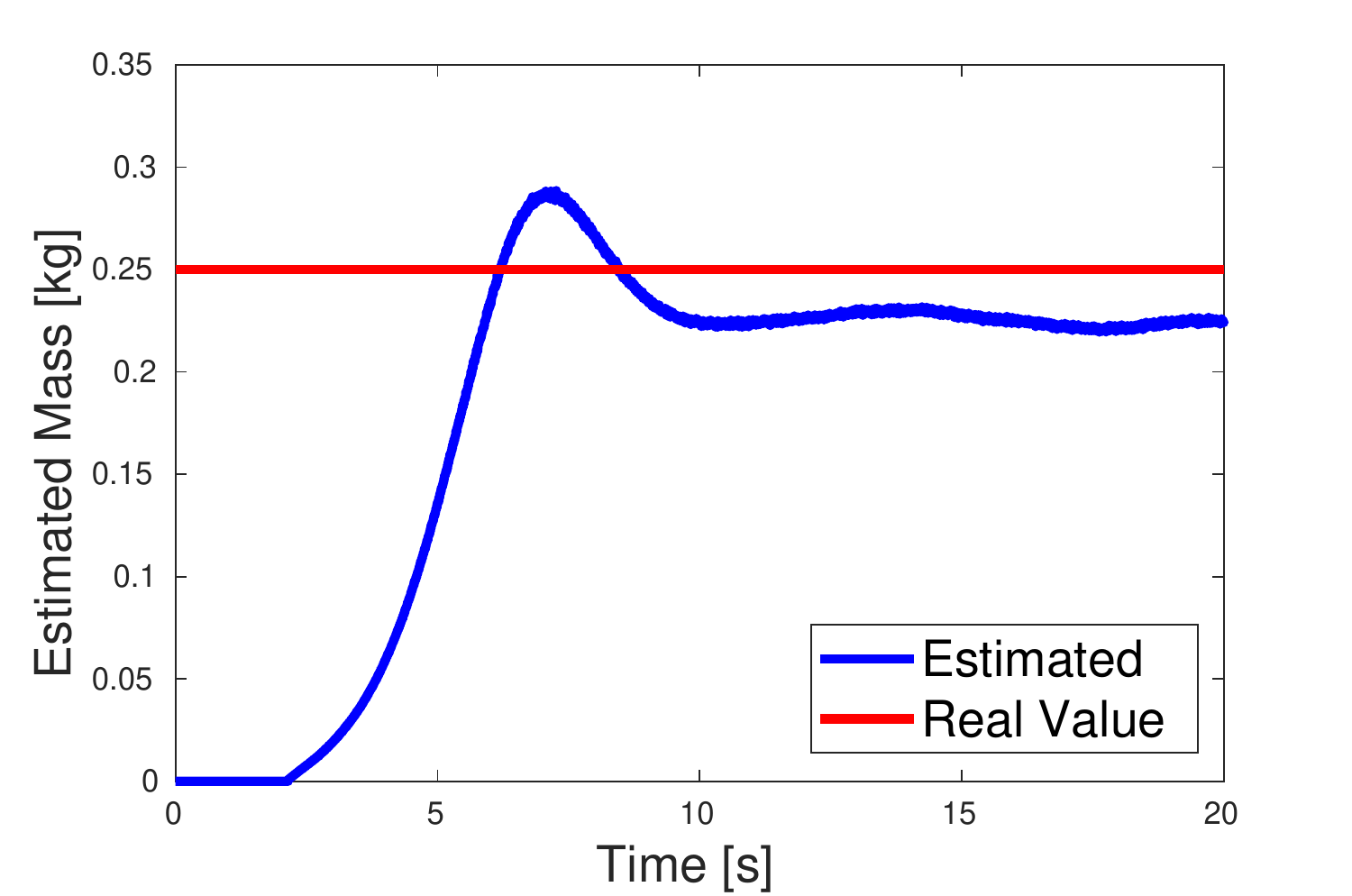}
    \caption{Overvalued 20\%}
    \label{fig:param_error3}
  \end{subfigure}
  \begin{subfigure}{0.3\textwidth}
    \centering
    \includegraphics[width=\textwidth]{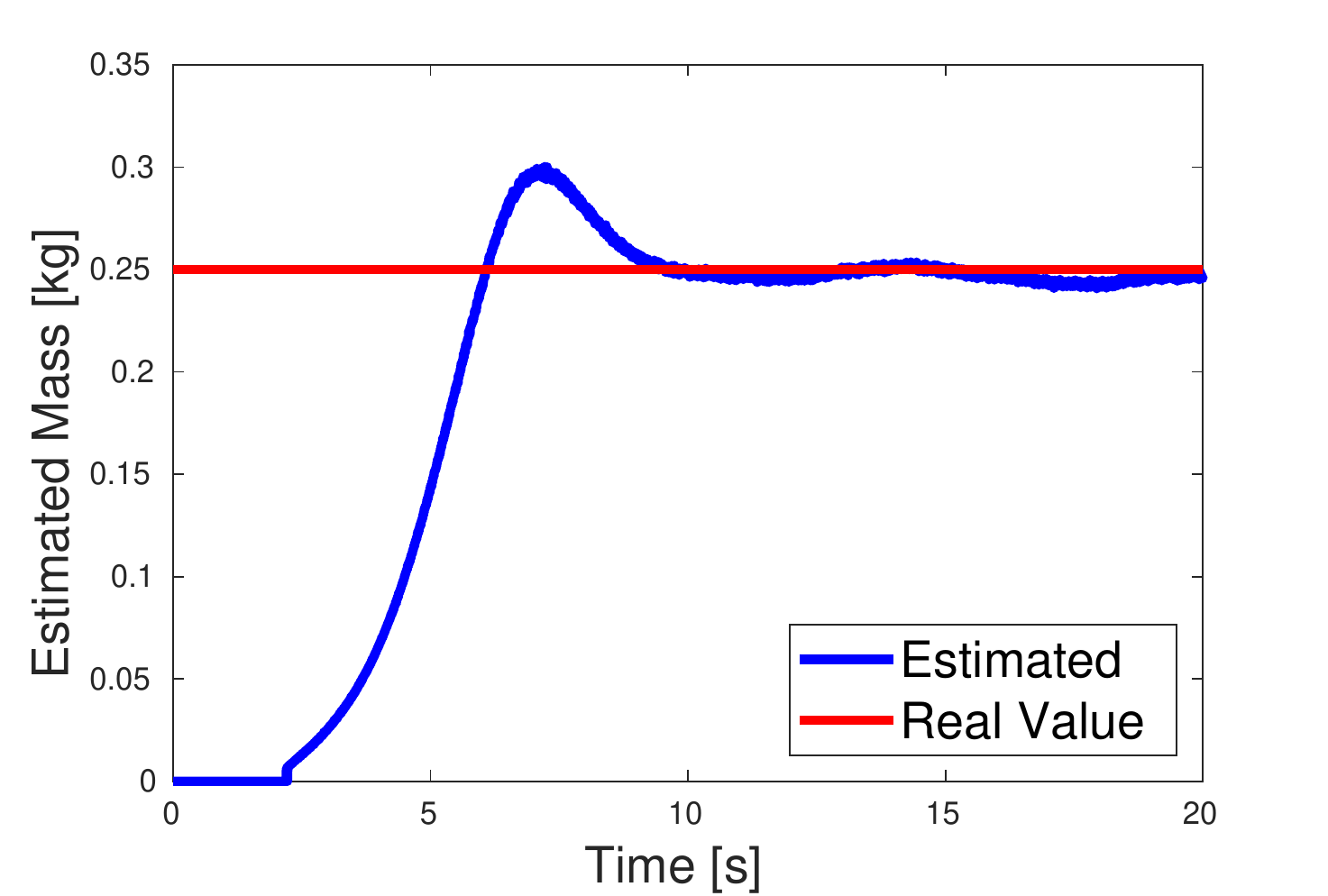}
    \caption{Correct values}
    \label{fig:param_error1}
  \end{subfigure}
  \caption{Effect of parameter uncertanties}
  \label{fig:ParamError}
\end{figure}

There are two possible causes of errors in the estimation: uncertanties in the physical parameters (link masses and inertia, joint frictions, etc) and the position of the object along the gripper. In this work it is assumed that the system is correctly identified, nevertheless, sometimes this assumption could not be held, specially in complex systems such as the 7-DOF manipulator. In order to analyze the effect of parameter uncertainties in the estimation, we ran simulations with wrong parameter values. That is, the term $f(x_1, \hat{x}_2, u)$ in (\ref{eq:observer1})-(\ref{eq:observer2}) is calculated with the wrong parameters but measurements $q$ come from the system simulated with the correct values. Figure \ref{fig:param_error2} shows the estimation results using link masses and inertia with values 20\% smaller than the nominal ones. As it can be seen, the system overestimates the mass value. The torque exerted on each joint is caused by the link mass and the manipulated object. If such link mass is smaller, then the exerted torque will be attributed to the manipulated object. The contrary effect happens when the physical parameters are overestimated, as shown in figure \ref{fig:param_error3}.

\begin{figure}[h!]
  \centering
  \begin{subfigure}{0.32\textwidth}
    \centering
    \includegraphics[width=\textwidth]{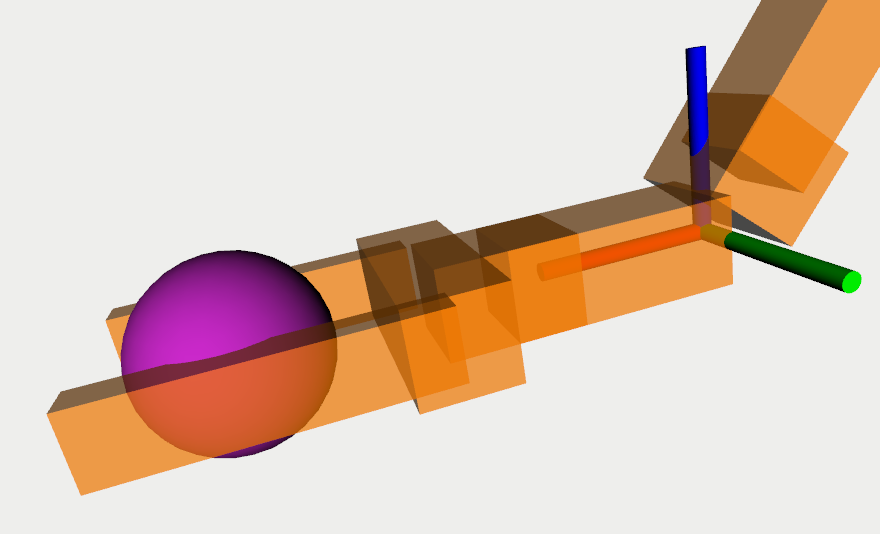}
    \includegraphics[width=\textwidth]{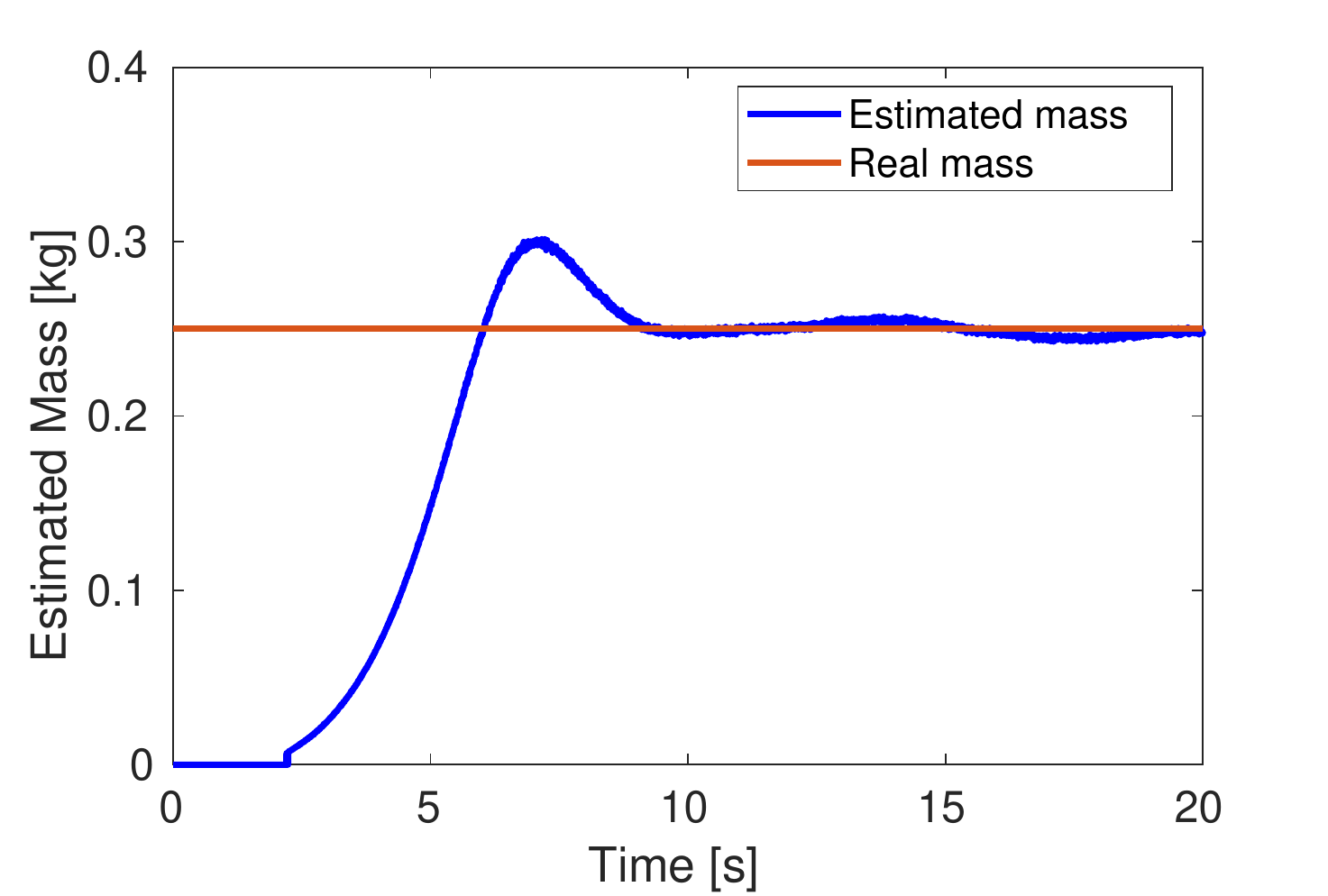}
    \caption{$l_4=0.21$ m}
    \label{fig:obj_position1}
  \end{subfigure}
  \begin{subfigure}{0.32\textwidth}
    \centering
    \includegraphics[width=\textwidth]{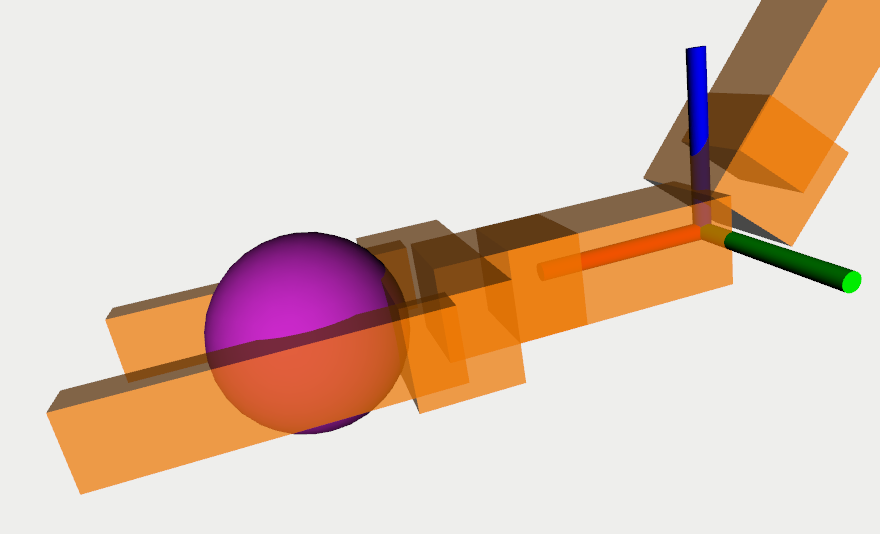}
    \includegraphics[width=\textwidth]{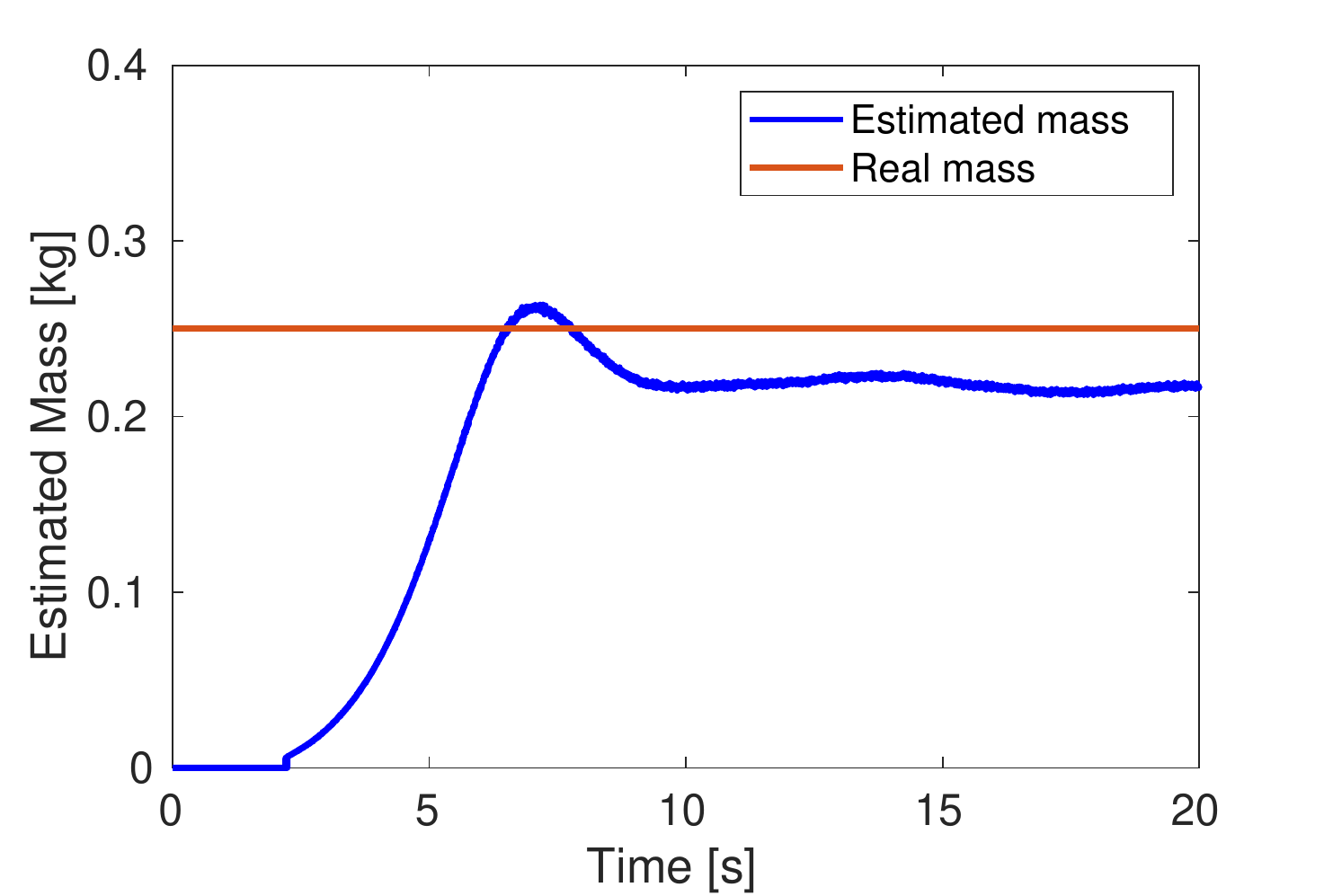}
    \caption{$l_4=0.18$ m}
    \label{fig:obj_position2}
  \end{subfigure}
  \begin{subfigure}{0.32\textwidth}
    \centering
    \includegraphics[width=\textwidth]{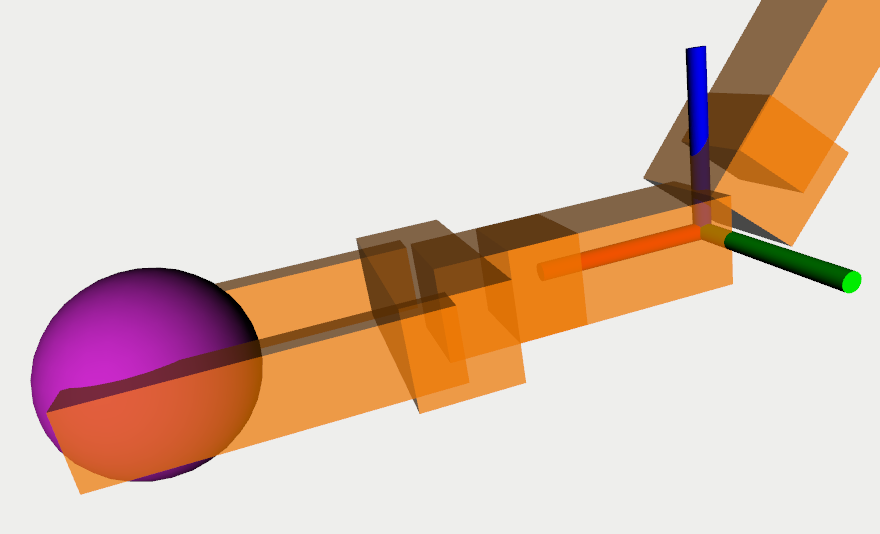}
    \includegraphics[width=\textwidth]{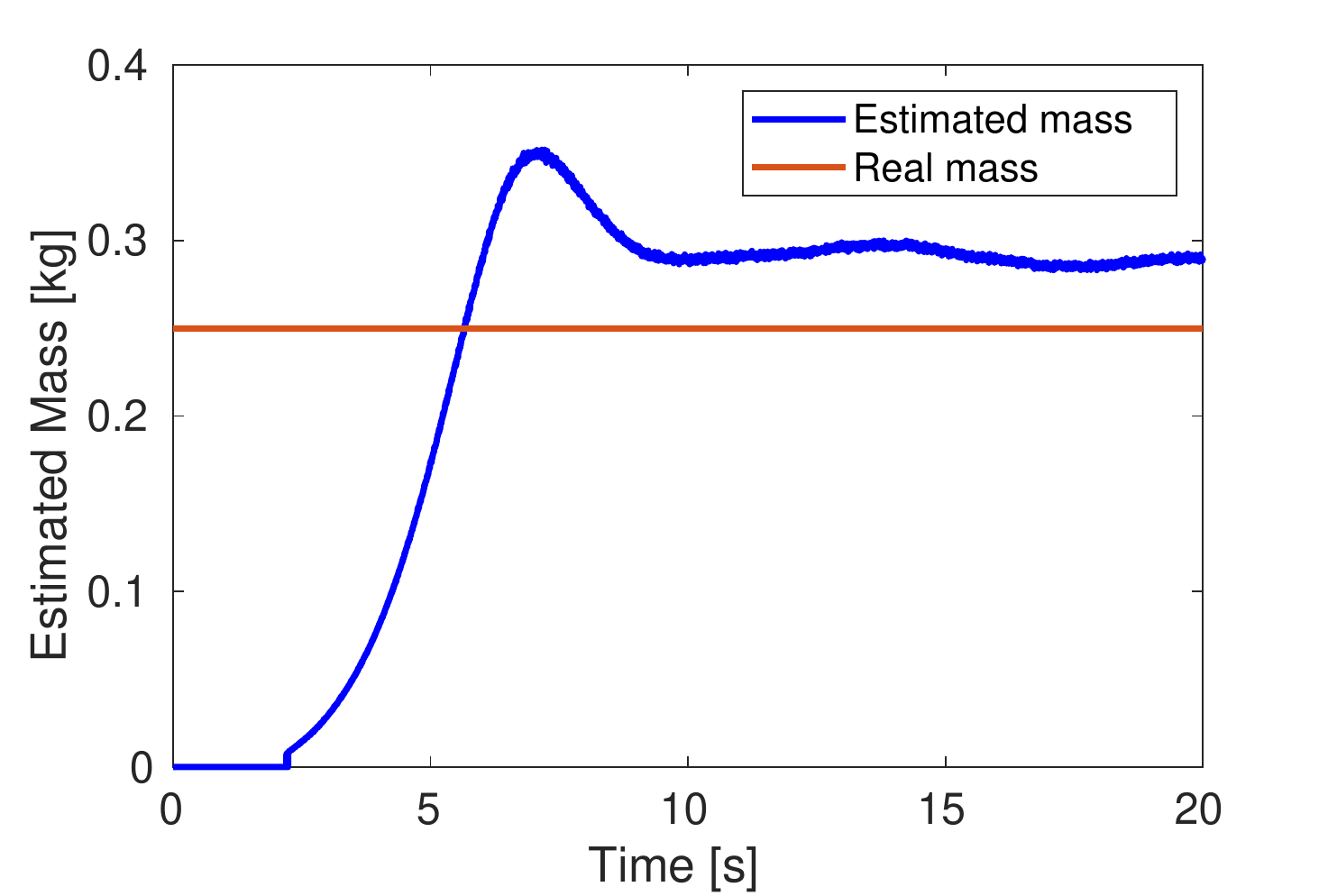}
    \caption{$l_4=0.24$ m}
    \label{fig:obj_position3}
  \end{subfigure}
  \caption{Effect of object position along gripper in mass estimation}
  \label{fig:ObjPosition}
\end{figure}
According to equation (\ref{eq:mass}), we need to know the distance $l_4$ for a correct mass estimation. This distance depends on the object position along the gripper. We always assume that the object is gripped in the center of the end effector, nevertheless, there could be a small error in this assumption. The error in the estimated mass will be proportional to the error in the estimation of $l_4$. Figure \ref{fig:ObjPosition} shows the estimated values for three different values of $l_4$. As it can be seen, the greater the distance $l_4$, the greater is the torque exerted on the joint with the same object mass and a greater value of mass is estimated. At this stage of the project, we don't have a way to determine the position of the object along the gripper, nevertheless, for the goal application of this project, the errors in estimation are tolerable.

\section{Reusability}
\label{sec:reusability}
To reproduce the results presented in this work, the following rough steps should be followed:
\begin{enumerate}
\item Get the URDF file of the manipulator and import it using the Simulink Robotics Toolbox. If the full robot URDF is available, the manipulator part should be isolated in order to make calculation faster.
\item Identify the joint with the following features:
  \begin{itemize}
  \item The joint must have an axis of rotation perpendicular (or nearly perpendicular) to the gravity vector. 
  \item The joint should be the nearest one to the end effector with the previous feature. 
  \end{itemize}
  And get the distance from the previous joint to the center of the end effector.
\item Implement SMO and use equation (\ref{eq:mass}) to estimate object mass. 
\end{enumerate}

To illustrate these steps, we will reproduce the results using the description of Katana Manipulator (\url{https://wiki.ros.org/katana_driver}).

\subsubsection*{Get and import the URDF}
\begin{figure}
  \centering
  \begin{subfigure}[b]{\textwidth}
    \centering
    \includegraphics[width=\textwidth]{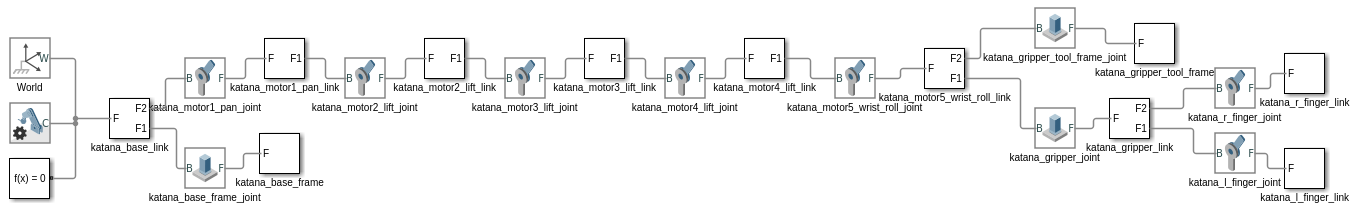}
    \caption{Imported Simscape Multibody model.}
    \label{fig:KatanaSimulinkA}
  \end{subfigure}
  \begin{subfigure}[b]{\textwidth}
    \centering
    \includegraphics[width=\textwidth]{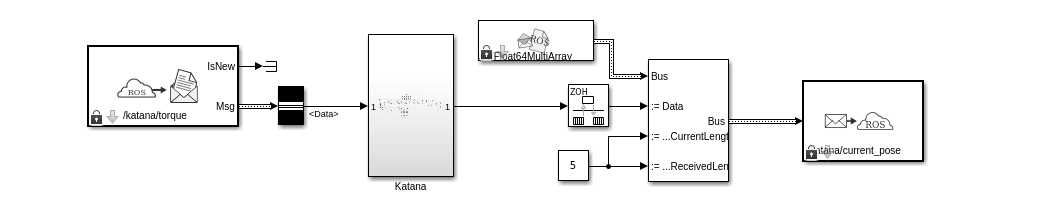}
    \caption{Manipulator with the corresponding publishers and subscribers.}
    \label{fig:KatanaSimulinkB}
  \end{subfigure}
  \begin{subfigure}[b]{0.65\textwidth}
    \centering
    \includegraphics[width=\textwidth]{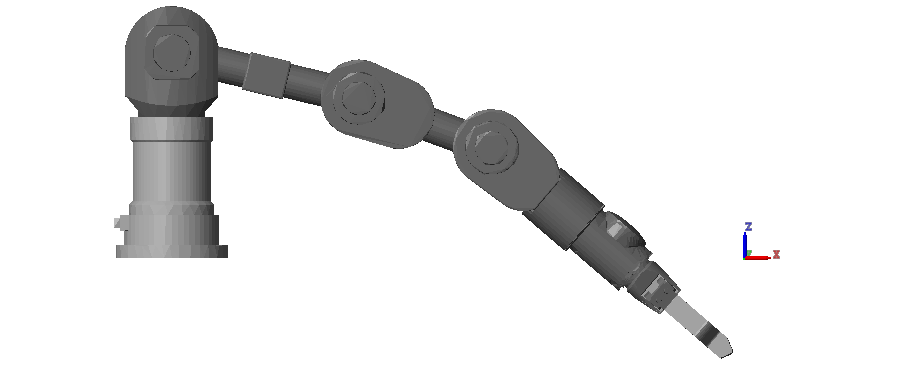}
    \caption{Manipulator as shown in the Mechanics Explorer.}
    \label{fig:KatanaSimulinkC}
  \end{subfigure}
  \caption{Importing Katana URDF to Simulink}
  \label{fig:KatanaSimulink}
\end{figure}

We imported the URDF from the GitHub Katana repository. Most robot descriptions are given using \texttt{xacro} formats, but the ROS Xacro package can be used to get the corresponding URDF, since Robotics System Toolbox and Simscape can only import URDF files. We used Simscape to import the URDF and we added the corresponding ROS publisher and subscriber for measured joint positions and torques respectively. Figure \ref{fig:KatanaSimulink} shows the manipulator as imported by Simscape, with ROS publishers and subscribers and the system as displayed in the Mechanics Explorer. 

\subsubsection*{Chosing the correct joint to estimate mass}
In this work we used the wrist pitch of our manipulator to estimate object mass. Using the observer (\ref{eq:observer1})-(\ref{eq:observer2}), we reconstructed the fault signals $\phi(x_1, x_2, u)$, after that, using equation (\ref{eq:taus}) we get the torques exerted on each joint. Finally, using torque on joint 6 (wrist pitch) and equation (\ref{eq:mass}), we get the estimated mass of the manipulated object. The joint of the wrist pitch fulfills the two features previously mentioned: it is the nearest joint to the end effector whose rotation axis is perpendicular to the gravity vector.

\begin{figure}
  \centering
  \includegraphics[width=0.7\textwidth]{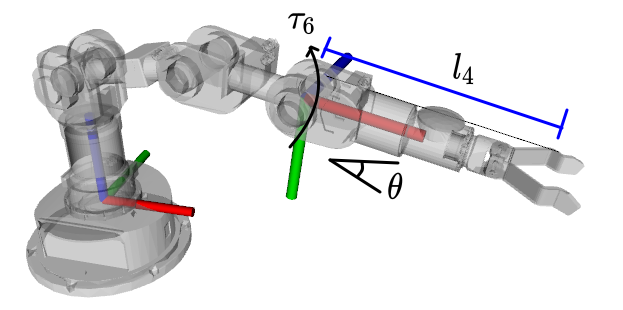}
  \caption{Katana equivalent variables for mass estimation}
  \label{fig:katana_joints}
\end{figure}

In the case of the Katana manipulator, the joint with these features is \texttt{katana\_motor4\_lift\_link} and thus, the fault torque associated to this joint will be used in equation (\ref{eq:mass}). Figure \ref{fig:katana_joints} shows the equivalent variables in the Katana Manipulator for mass estimation. Angle $\theta$ can be calculated using the Euler Angles of the rotation from \texttt{katana\_base\_frame} to \texttt{katana\_motor4\_lift\_link}. Robotics System Toolbox provides the needed blocks to calculate this angle. Distance $l_4$ can be obtained from URDF or by directly measuring it. 

\subsubsection*{SMO implementation and mass estimation}
\begin{figure}
  \centering
  \includegraphics[width=0.5\textwidth]{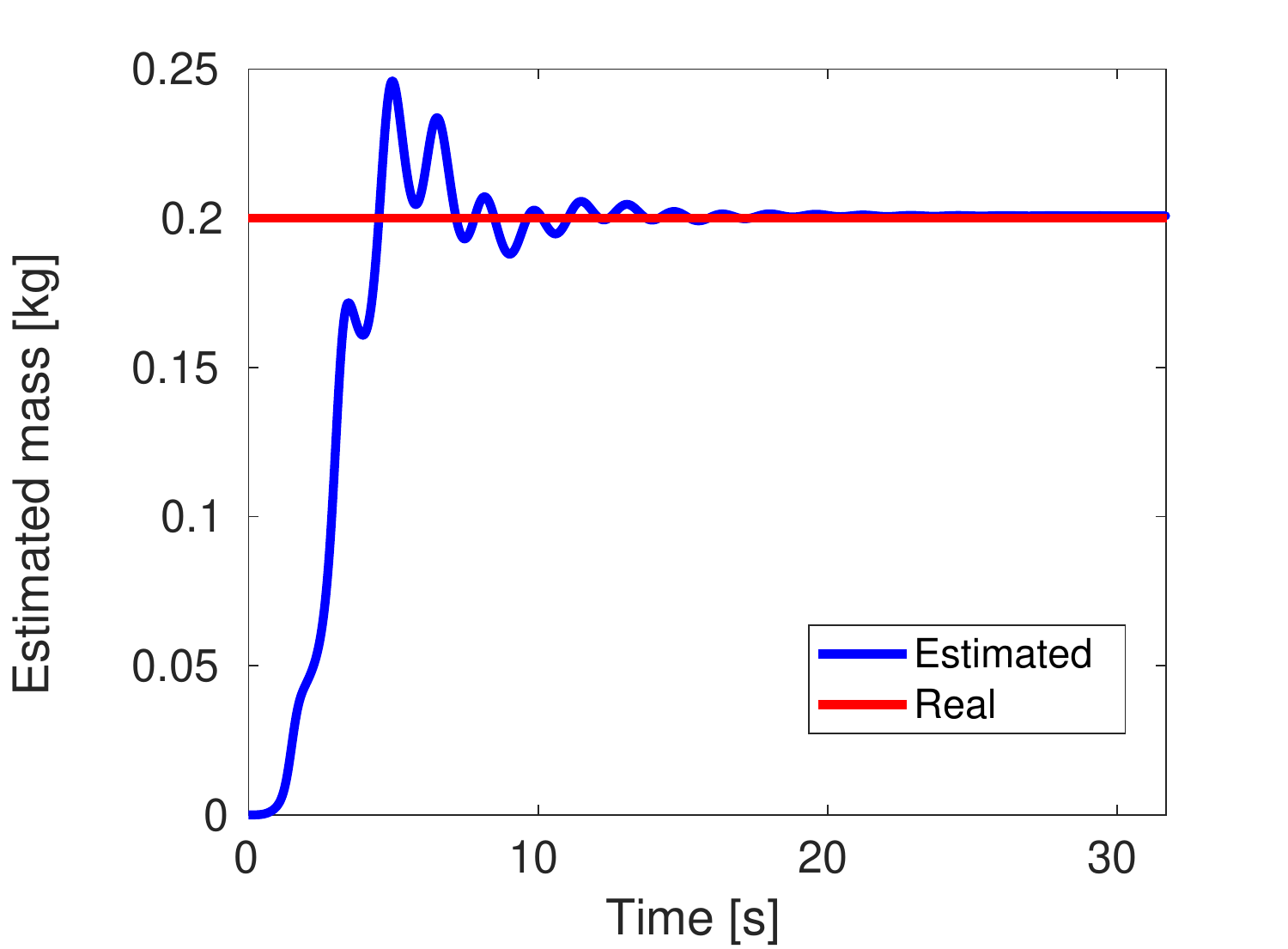}
  \caption{Mass estimation results with the Katana Manipulator}
  \label{fig:katana_results}
\end{figure}

Once we identified the ideal joint for mass estimation, we imported the URDF file to get a Rigid Body Tree using the \texttt{importrobot} function of the Robotics System Toolbox. We implemented the SMO and mass estimator using the Forward Dynamics and Mass Matrix block as shown in figure \ref{fig:SimulinkMassEst} (see explanation in section \ref{sec:implementation}). We simulated an object with a mass of 200 g. To simplify simulations, we did not implement a control but we used only a constant torque applied to the joints. Figure \ref{fig:katana_results} shows the resulting estimation. 

All files and instructions necessary to reproduce the results presented in this work can be found in\\ \url{https://github.com/RobotJustina/RCF-MathWorks-2020-14/}. 

\section{Conclusions}
\label{sec:conclusions}
In this work we presented a method for estimating the mass of the object being manipulated in a domestic service robot. The estimation is based on model based fault reconstruction techniques using sliding mode observer. The implementation of model-based algorithms could be very complicated in complex systems such as the 7 DOF manipulator used in this proposal; nevertheless, the use of Simulink, Robotics System Toolbox, Simscape and ROS Toolbox, which provide numerical solutions to the system dynamics, allowed the implementation of such model-based techniques. We presented simulation results where we show the effectiveness of our proposal. Also, we discussed the cases where this approach can lead to a wrong estimation. To show the possible reusability of our proposal, we replicated some of the results in a simulated Katana manipulator. 

Further tests should be made for an implementation in the real manipulator. For example, in this work we assumed that only linear friction is present in the motor, nevertheless, Dynamixel motors present also non linear frictions, which must be modeled and identified for a correct performance. Also, better controllers can be implemented to reach faster the steady state. A shorter settle time will result in a faster mass estimation. 

As a future work, different techniques can be implemented. Another model-based option is the use of the implemented Extended Kalman Filter. As shown in section \ref{sec:results}, the EKF estimated speeds have a steady state error due to the fault signal. This estimated speeds can also be used to estimate the manipulated object mass. On the other hand, model-free techniques can also be implemented, such as neural networks. Input torques and measured positions can be taken as input data to train a NN whose output is the estimated mass.

Finally, this proposal will be integrated with the rest of subsystems of our domestic service robot in the context of the Robocup@Home tests.

\bibliographystyle{abbrv}
\bibliography{References}
\end{document}